\documentclass{article} 
\PassOptionsToPackage{table}{xcolor}
\usepackage{iclr2027_conference,times}

\usepackage{amsmath,amsfonts,bm}

\def\eqref#1{equation~\ref{#1}}

\def\1{\bm{1}}

\DeclareMathAlphabet{\mathsfit}{\encodingdefault}{\sfdefault}{m}{sl}
\SetMathAlphabet{\mathsfit}{bold}{\encodingdefault}{\sfdefault}{bx}{n}

\usepackage{hyperref}
\usepackage{url}
\usepackage{graphicx}
\usepackage{booktabs}
\usepackage{xcolor}
\usepackage{wrapfig}

\usepackage{float}
\usepackage{xspace}
\newcommand{\method}{GAE\xspace}

\title{\method: Learning a Geometry-Native Latent Space for 3D-Consistent World Generation \vspace{0.4em}}

\author{
\begin{minipage}{1\textwidth}
\centering
\normalfont
{\large
\textbf{Jiahao Lu}$^{1*}$\quad
\textbf{Minghao Yin}$^{2,3*}$\quad
\textbf{Wenbo Hu}$^{2\dagger}$\quad
\textbf{Hengyu Liu}$^{4}$
\\[0.45em]
\textbf{Wang Zhao}$^{2}$\quad
\textbf{Sai-Kit Yeung}$^{1}$\quad
\textbf{Ying Shan}$^{2}$\quad
\textbf{Yuan Liu}$^{1\dagger}$
}
\\[0.8em]
{\small
$^{1}$The Hong Kong University of Science and Technology
\quad\textperiodcentered\quad
$^{2}$ARC Lab, Tencent IEG
\\[0.25em]
$^{3}$The University of Hong Kong
\quad\textperiodcentered\quad
$^{4}$The University of Texas at Austin
}
\\[0.65em]
{\footnotesize
$^{*}$Equal contribution
\qquad
$^{\dagger}$Corresponding authors
} 
\\[0.75em]
\href{https://jiah-cloud.github.io/GAE.github.io/}{%
  \textcolor{blue!65!black}{\textbf{Project Page}}}
\end{minipage}
}

\newcommand{\best}[1]{\cellcolor{blue!22}\textbf{#1}}
\newcommand{\snd}[1]{\cellcolor{blue!9}#1}

\iclrfinalcopy 

\begin{document}

\maketitle

\begin{figure}[!ht]
 \vspace{-3em}
    \centering
    \includegraphics[
        width=\linewidth,
        trim=0 0 0 0,
        clip
    ]{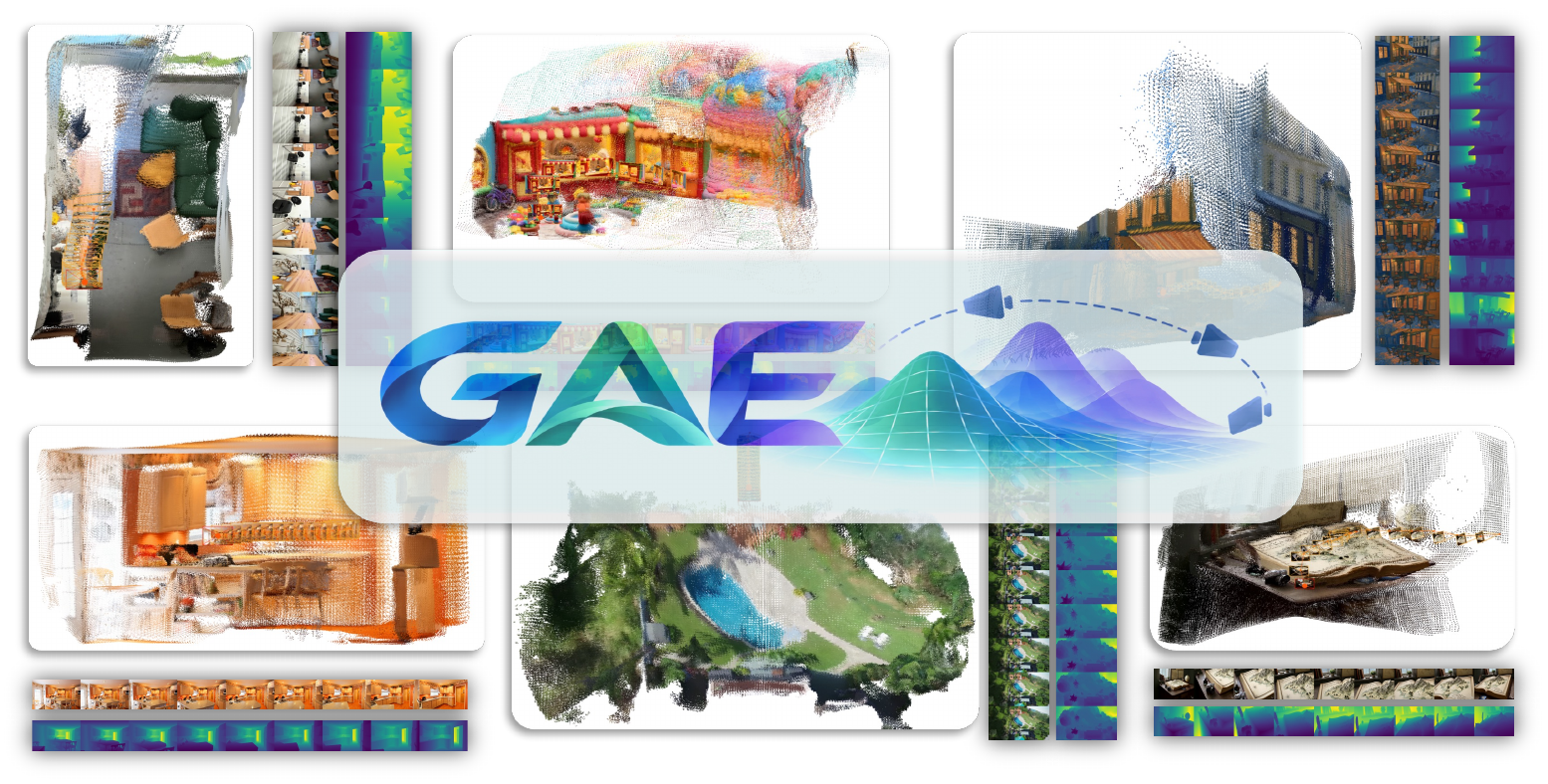}
    \vspace{-2em}
    \caption{
    \textbf{Geometry-native world generation.}
    Given a reference image, a camera trajectory, and optional text, \method
    jointly generates RGB views and 3D geometry from
    \emph{a shared, compact latent space}.
    Each example pairs generated RGB and depth sequences with the resulting
    3D point cloud.
    }
    \label{fig:teaser}
\end{figure}

\begin{abstract}

We present a compact geometry-native latent space as a shared foundation for perception and generation.
Visual generators can produce photorealistic frames without preserving a consistent 3D scene.
We argue that this is not only a modeling problem but also a representation problem: generators typically evolve appearance-centric latents, while perception models recover geometry in
a semantically rich space that encodes cross-view structure.
%
Rather than adding geometry as another output, we reparameterize a geometry foundation model's features into a compact latent space for generation.
We realize this shift with the \emph{geometry-native autoencoder} (\method), whose latent is jointly decodable to appearance, depth, cameras, and point maps.
With this state, a standard conditional flow supports diverse generation tasks.
In controlled comparisons that hold the generator and training protocol fixed, replacing the latent with \method improves both visual quality and independently measured 3D coherence: FVD falls by $12.7\%$ and $23.1\%$ on RealEstate10K and DL3DV, and camera-trajectory error is halved on RealEstate10K.
Together, these results show that the latent space is central to geometry-consistent generation and can serve as a shared interface between perception and generation.
\end{abstract}


\section{Introduction}
\label{sec:intro}

Photorealistic frames do not guarantee a coherent scene.
From a few real views of one scene, geometry foundation models can recover depth, cameras, and point maps that agree in a common 3D coordinate frame~\citep{wang2024dust3r,wang2025vggt}.
Video generators, however, can produce views whose recovered geometry drifts and camera trajectories stray from the requested path. 
\emph{What perception can read, generation still struggles to write.}
If video generation is to serve as a foundation for world models, its frames must describe a persistent scene.
We trace this requirement upstream, to the latent
space
the generator is trained to evolve.

A latent state does more than compress the output.
It shapes which structure is directly available to the generator and which must be inferred from appearance or introduced elsewhere.
A pixel VAE preserves appearance~\citep{kingma2014vae,rombach2022ldm}, while a representation autoencoder organizes semantic content~\citep{zheng2026rae,raev22026}, but neither makes depth, camera geometry, and cross-view relations natively readable.
The usual remedy is to add geometry alongside an appearance-native latent, through camera controls, aligned representations, or geometry-aware post-training~\citep{he2025cameractrl,gen3r2026,vggrpo2026}.
\emph{We argue that perception and generation should instead share a geometry-native latent space.}

Geometry foundation models offer a natural starting point for such a latent~\citep{wang2024dust3r,wang2025vggt,lin2026da3}.
%
Yet their geometry decoders typically rely on a hierarchy of features rather than a single latent representation.
Different levels provide complementary cues, including local
detail, correspondence, semantic context, and cross-view structure.
This division of labor is effective for perception but awkward for generation:
modeling the full hierarchy introduces multiple coupled generative states,
whereas selecting one level discards information expected by the geometry
decoder. GLD~\citep{gld2026} illustrates the former approach by using a
cascade of flow models, but the underlying mismatch applies more broadly to
models that decode geometry from multi-level features.
We use DA3~\citep{lin2026da3} as a controlled case study to make this broader
perceptual--generative mismatch concrete: no single level simultaneously
preserves the semantic, spatial, and cross-view structure needed for
geometry-aware generation while providing a smooth, well-conditioned space for
generative transport.
Earlier features retain stronger scene structure but are harder to model continuously, whereas later features become smoother at the cost of substantial geometric and semantic information.
Moreover, individual levels remain highly redundant and anisotropic: despite having 3,072 channels, they span only about 11 effective dimensions, with condition numbers ranging from $10^{8}$ to $10^{16}$.
This leaves most channels nearly inactive and concentrates useful information along a small number of directions at vastly different scales.
Therefore, the geometry is already present, but its raw parameterization is poorly matched to generation.


To reparameterize this hierarchy for generation, we introduce the \emph{geometry-native autoencoder} (\method).
%
Rather than directly modeling the feature hierarchy, as in GLD~\citep{gld2026},
we argue that a generative reparameterization should satisfy three principles:
(1) \emph{Preserve} the geometry already encoded by the perception model; (2) \emph{Unify} appearance and geometry within a single compact representation; and (3) \emph{Organize} that representation
for smooth generative transport.
To preserve geometry, \method places a learned bottleneck between DA3's frozen encoder and geometry head, compressing all four feature levels into a single latent and reconstructing the full hierarchy in one pass. Crucially, the original geometry head remains frozen when decoding depth, camera rays, and point maps, preventing the decoder from adapting around information
lost during compression.
To unify geometry and appearance, a learned RGB head decodes appearance from the same bottleneck, forcing both to coexist in a shared latent.
These readouts constrain \emph{what} information the latent preserves, but not \emph{how}
it
is organized for flow modeling.
%
We therefore organize
the bottleneck at two complementary levels: individual tokens and relations among them. Token-wise alignment to co-located C-RADIO features~\citep{heinrich2025radio} improves transport smoothness and semantic organization, but on its own sharply degrades pairwise spatial structure.
Matching
pairwise similarities in the posterior to those in DINOv2 features~\citep{oquab2024dinov2} restores this relational geometry while retaining the token-level gains.
%
\method thus preserves geometry, unifies it with appearance in a compact latent, and organizes that latent natively for generation.

Using this latent, we train a unified generator based on a standard DiT-style conditional flow~\citep{lipman2023flow}.
The same model supports text-to-image generation, camera-controlled video, and reference-conditioned novel-view synthesis.
Condition dropout enables this breadth by exposing the model to different combinations of text, metric Pl\"ucker rays~\citep{scope2026}, and reference views.
Across regimes, the model jointly generates all target-view latents, which decode natively to both RGB and geometry.
Perception and generation therefore share the same latent space.

To isolate the effect of the latent space, we compare \method with pixel, semantic, and raw geometry latents~\citep{rombach2022ldm,raev22026,lin2026da3} using matched flow models and training protocols.
%
\method reduces FVD by $12.7\%$ on RealEstate10K~\citep{zhou2018re10k} and $23.1\%$ on DL3DV~\citep{ling2024dl3dv} relative to the strongest competing latent.
Independent evaluation of the generated views also shows stronger 3D consistency, with camera-trajectory error roughly halved on RealEstate10K.
%
Together, these results establish a compact geometry-native latent space as a shared foundation for perception and generation.

\section{Related Work}

\paragraph{Generative representation spaces.}
Visual generation has separated perceptual compression from generative modeling through variational and vector-quantized tokenizers, with latent diffusion establishing compact codes as the standard generative state~\citep{kingma2014vae,oord2017vqvae,esser2021taming,rombach2022ldm}.
Representation autoencoders (RAEs) instead use frozen visual representations as the encoder state, with later work scaling the idea to text-to-image generation and multi-layer features~\citep{zheng2026rae,tong2026scalerae,raev22026}.
REPA aligns denoiser features rather than changing the output state, while latent-diffusability studies show that reconstruction alone does not determine generation quality~\citep{yu2025repa,diffusingrightspace2026}.
These methods organize generation around appearance or semantics.
\method instead asks for a compact state that remains natively readable by a geometry foundation model.

\paragraph{Geometry-aware visual generation.}
Scene-based novel-view synthesis reconstructs radiance fields or Gaussian primitives, while generative methods use pose-conditioned diffusion to synthesize unobserved content from text or sparse views~\citep{mildenhall2020nerf,kerbl2023gaussians,liu2023zero123,shi2024mvdream,liu2024syncdreamer,gao2024cat3d,yu2024polyoculus,yu2025viewcrafter}.
Camera-controlled video similarly introduces trajectories through conditioning or positional encoding~\citep{wang2024motionctrl,he2025cameractrl,scope2026}.
Joint-output methods generate RGB together with depth or normals~\citep{stan2023ldm3d,krishnan2025orchid,kwon2025jointdit,guizilini2025mvgd}.
Other approaches align video features to geometry representations, condition on geometry features, couple geometry and appearance latents, or post-train with geometric rewards~\citep{wu2026geometryforcing,wan2026geoworld,gen3r2026,mi2026one4d,xiang2026videoweave,vggrpo2026}.
More recently, 3DRAE~\citep{wei2026any} aggregates frozen 2D representations into fixed-length, view-decoupled 3D latent tokens and performs diffusion directly in this scene-level latent space. 
Across these generative designs, geometry augments an appearance-led state as a control, modality, auxiliary representation, or objective.
\method instead derives the sole evolving latent from a geometry foundation model rather than adding geometry to an appearance-native state.

\paragraph{Geometry foundation models as generative states.}
Geometry foundation models have progressed from paired point-map regression and matching to many-view, persistent, reference-free, and scaled static/dynamic reconstruction~\citep{wang2024dust3r,leroy2024mast3r,lu2025align3r,wang2025cut3r,wang2025vggt,lin2026da3,wang2026pi3,wang2026vggtomega}.
Pretrained features have also become predictive states for planning, geometry forecasting, and stochastic world modeling~\citep{zhou2025dinowm,vggtworld2026,flowwm2026}.
The closest visual-generation precedents directly model geometry-foundation features: GLD cascades selected DA3 or VGGT levels and propagates the remaining hierarchy~\citep{gld2026}, latent Riemannian flow matching jointly evolves VGGT's four normalized levels on their product manifold~\citep{weijler2026rfm}, and OneWorld~\citep{gao2026oneworld} augments Pi3 geometry features with appearance and semantic information before modeling the resulting unified 3D representation.
These approaches operate directly on high-dimensional backbone representations or adapt them for use as high-dimensional generative states.
\method instead learns one compact Euclidean reparameterization of the full hierarchy, evolves it with a standard flow model, and reconstructs the hierarchy for frozen geometry readout and learned RGB decoding.\

\section{Method}
\label{sec:method}

\begin{figure}[t]
    \centering
    \includegraphics[width=\linewidth]{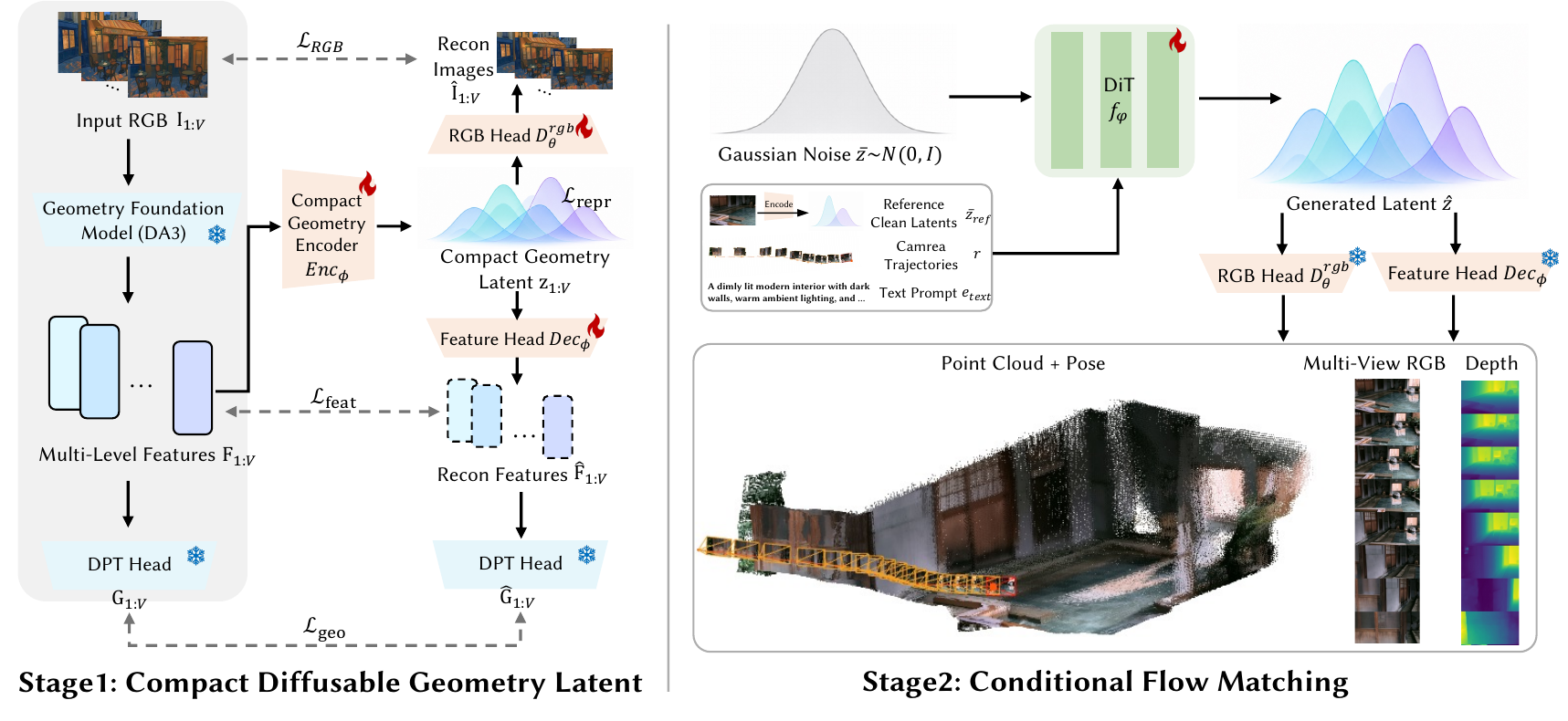}
    \vspace{-2em}
    \caption{\textbf{Overview of Geometry-Native Autoencoder (\method).}
    \emph{Stage~1 (codec training)} compresses frozen multi-level geometry
    foundation features into a compact, diffusable geometry latent; the feature
    decoder rebuilds the full hierarchy for frozen geometry readout, while a
    separate learned head decodes RGB. \emph{Stage~2 (flow training)} performs
    conditional flow matching only over generated-view latents, while clean
    reference latents, camera rays, and text act as controls.}
    \label{fig:ngd_pipeline}
\end{figure}

Geometry-Native Autoencoder (\method) places the 3D inductive bias in the generated
state itself. The method trains in two stages, shown in
Fig.~\ref{fig:ngd_pipeline}: Stage~1 trains a codec that turns the multi-level
features of a frozen geometry backbone into one compact latent decodable to both
RGB and geometry, and Stage~2 freezes that codec and trains a conditional flow
model in its standardized latent space.
The two stages are summarized by:
\begin{equation}
\underbrace{
\mathbf{I}_{1:V}
\xrightarrow{\;\mathcal{E},\,\mathcal{P}\;}
\mathbf{X}_{1:V}
\xrightarrow{\;\mathrm{Enc}_{\phi}\;}
\mathbf{z}_{1:V}
\xrightarrow{\;\mathrm{Dec}_{\phi}\;}
\hat{\mathbf{F}}_{1:V}^{0:3}
}_{\text{Stage 1: train the codec }\mathcal{A}_{\phi}},
\qquad
\underbrace{
\bar{\mathbf{z}}^{(1)}_{1:V}
\xrightarrow{\;\mathrm{Flow}_{\psi}(\cdot\mid\mathcal{C})\;}
\hat{\bar{\mathbf{z}}}_{1:V}
\xrightarrow{\;\mathrm{Std}^{-1}\;}
\hat{\mathbf{z}}_{1:V}
}_{\text{Stage 2: train the flow model }\mathbf{f}_{\psi}}.
\label{eq:ngd_pipeline}
\end{equation}
Here $\mathcal{E}$ is the frozen DA3 encoder and $\mathcal{P}$ normalizes and
fuses its four feature levels. Stage~1 trains $\mathrm{Enc}_{\phi}$ and
$\mathrm{Dec}_{\phi}$ to compress $\mathbf{X}_{1:V}$ and recover the four-level
hierarchy $\hat{\mathbf{F}}_{1:V}^{0:3}$; once frozen, the codec's posterior mean
standardized by $\mathrm{Std}$ supplies the flow targets
$\bar{\mathbf{z}}^{\mathrm{gt}}_{1:V}$. Flow sampling starts from
$\bar{\mathbf{z}}^{(1)}\sim\mathcal{N}(\mathbf{0},\mathbf{I})$.
The condition set $\mathcal{C}$ may contain text, camera rays, and clean latent
tokens encoded from reference views. Only the target-view latents evolve as ODE
states; all other signals remain fixed controls.
Both outputs are read from the sampled latent:
\begin{equation}
    \hat{\mathbf{I}}_{1:V}
    = \mathcal{D}_{\theta}^{\mathrm{rgb}}(\hat{\mathbf{z}}_{1:V}),
    \qquad
    \hat{\mathbf{G}}_{1:V}
    = \mathcal{H}_{\mathrm{DPT}}
      \!\left(\mathrm{Dec}_{\phi}(\hat{\mathbf{z}}_{1:V})\right).
    \label{eq:ngd_readout}
\end{equation}
The trainable codec
$\mathcal{A}_{\phi}=(\mathrm{Enc}_{\phi},\mathrm{Dec}_{\phi})$ reconstructs the
full DA3 hierarchy~\cite{lin2026da3}, which the frozen geometry head
$\mathcal{H}_{\mathrm{DPT}}$ reads as depth, rays, and point maps. A separate
learned head $\mathcal{D}_{\theta}^{\mathrm{rgb}}$ renders RGB from the same
latent. Thus, unlike a pixel VAE, the state remains geometry-decodable; unlike
raw-feature diffusion, it is compact enough to be modeled by a single flow
network. We use $\mathbf{z}$ for the codec latent, $\bar{\mathbf{z}}$ for its
standardized form used by flow matching, and
$\bar{\mathbf{z}}^{\mathrm{ref}}$ for clean reference evidence.

\subsection{Geometry-native codec}
\label{sec:method_codec}
\label{sec:method_autoencoder}

The DA3 hierarchy presents two unsatisfactory choices for generation. Modeling
all four raw levels preserves the information needed for geometry decoding but
entails modeling a high-dimensional, coupled feature hierarchy; selecting one
level reduces the state size but discards complementary information and retains the poor
conditioning identified in Section~\ref{sec:exp_latent}. \method instead learns the
state between a frozen DA3 encoder and its frozen geometry head: it fuses the
hierarchy, compresses it into one spatial latent, and reconstructs all levels for
joint RGB and geometry decoding.

\paragraph{From a feature hierarchy to one state.}
Given $V$ input views, the frozen DA3 encoder $\mathcal{E}$ produces the
four-level hierarchy
$\mathbf{F}_v^{0:3}:=\{\mathbf{f}_{v,\ell}\}_{\ell=0}^{3}$, where
$\mathbf{f}_{v,\ell}\in\mathbb{R}^{N\times C_\ell}$ contains the patch tokens
for view $v$ at level $\ell$ and $N=h_pw_p$. The shallow level preserves fine
detail and cross-view correspondence, while the deeper levels provide
complementary inputs required by the frozen geometry decoder. This distinction
does not imply stronger semantic or spatial structure at greater depth; our
diagnostics show the opposite for L3 (Table~\ref{tab:latent_diagnostics}).

Before fusion, we normalize each level using fixed, per-channel training-set
statistics $\mathbf{m}_\ell$ and $\mathbf{s}_\ell$. We then reshape the patch
tokens to their spatial grids and concatenate them along the channel dimension:
\begin{equation}
    \mathbf{X}_v = \mathcal{P}(\mathbf{F}_v^{0:3}) :=
    \big\|_{\ell=0}^{3}
    \mathrm{reshape}\!\left(
        \frac{\mathbf{f}_{v,\ell}-\mathbf{m}_\ell}
             {\mathbf{s}_\ell+\epsilon}
    \right)
    \in\mathbb{R}^{C_x\times h_p\times w_p},
    \qquad C_x=\sum_{\ell=0}^{3}C_\ell ,
    \label{eq:feat_fusion}
\end{equation}
where $\|$ denotes channel-wise concatenation. Thus, $\mathcal{P}$ is a fixed
operator that converts the four DA3 token matrices into the single fused tensor
$\mathbf{X}_v$. Level-wise normalization prevents high-variance channels from
dominating this representation.

\paragraph{Compact geometry codec.}
The feature codec
$\mathcal{A}_\phi:=(\mathrm{Enc}_\phi,\mathrm{Dec}_\phi)$ compresses the
fused tensor into a grid-shaped latent
$\mathbf{z}_v\in\mathbb{R}^{C_z\times h_p\times w_p}$ and reconstructs the
four-level hierarchy:
\begin{align}
    (\boldsymbol{\mu}_v,\log\boldsymbol{\sigma}_v^2)
        &= \mathrm{Enc}_\phi(\mathbf{X}_v), \\
    \mathbf{z}_v
        &= \boldsymbol{\mu}_v+
           \boldsymbol{\epsilon}_v\odot
           \exp\!\left(\tfrac{1}{2}\log\boldsymbol{\sigma}_v^2\right),
           \qquad
           \boldsymbol{\epsilon}_v\sim\mathcal{N}(\mathbf{0},\mathbf{I}),
           \label{eq:vae_reparam} \\
    \hat{\mathbf{F}}_v^{0:3}
        &= \mathrm{Dec}_\phi(\mathbf{z}_v).
        \label{eq:vae_decode}
\end{align}
\method preserves the DA3 patch grid and compresses only along channels, yielding
$C_z\in\{64,128\}$ rather than the 3,072 channels of a single raw DA3 level. The
encoder and decoder use lightweight convolutional pyramids with spatial
self-attention, retaining the DA3 patch layout while allowing long-range mixing.
We sample $\mathbf{z}_v$ during codec training, but use the posterior mean
$\boldsymbol{\mu}_v$ deterministically for flow training and inference. A small
KL weight regularizes the bottleneck without forcing it toward a heavily
regularized generative-VAE regime.

\paragraph{Joint RGB and geometry decoding.}
In the notation above, $\mathrm{Dec}_\phi$ includes reconstruction of the
normalized fused tensor followed by the fixed inverse split and denormalization,
and therefore returns
$\hat{\mathbf{F}}_v^{0:3}=\{\hat{\mathbf{f}}_{v,\ell}\}_{\ell=0}^{3}$.
A separate learned RGB head renders appearance directly from the compact latent,
whereas the original frozen DA3 dense-prediction transformer (DPT) head reads
geometry from the reconstructed hierarchy:
\begin{equation}
    \hat{\mathbf{I}}_v =
    \mathcal{D}^{\mathrm{rgb}}_\theta(\mathbf{z}_v),
    \qquad
    \hat{\mathbf{G}}_v =
    \mathcal{H}_{\mathrm{DPT}}(\hat{\mathbf{F}}_v^{0:3})
    = \mathcal{H}_{\mathrm{DPT}}
      \!\left(\mathrm{Dec}_\phi(\mathbf{z}_v)\right),
    \label{eq:dpt_decode}
\end{equation}
where $\hat{\mathbf{G}}_v$ contains depth, rays, and derived point maps.
Consequently, RGB and geometry are two readouts of the same compact state.
Keeping $\mathcal{H}_{\mathrm{DPT}}$ frozen prevents the geometry head from
adapting to information lost by the codec: successful geometry reconstruction
must remain readable by the original DA3 head. Unlike our raw single-layer
baselines, \method reconstructs the entire hierarchy directly and requires no
backbone propagation between the generated state and geometry decoding.

\paragraph{Codec objective.}
We first require the bottleneck to reconstruct both the hierarchy and its
downstream readouts:
\begin{equation}
    \mathcal{L}_{\mathrm{codec}} =
    \mathcal{L}_{\mathrm{feat}}
    + \lambda_{\mathrm{kl}}\mathcal{L}_{\mathrm{kl}}
    + \mathcal{L}_{\mathrm{rgb}}
    + \mathcal{L}_{\mathrm{geo}}
    + \mathcal{L}_{\mathrm{repr}}.
    \label{eq:codec_loss}
\end{equation}
Here $\mathcal{L}_{\mathrm{feat}}$ is a weighted level-wise feature
reconstruction loss, $\mathcal{L}_{\mathrm{rgb}}$ combines pixel and perceptual
reconstruction terms, and $\mathcal{L}_{\mathrm{geo}}$ supervises depth and ray
outputs. The geometry targets are pseudo-targets obtained by applying the frozen
DA3 head to the original features, rather than direct ground truth. The final term $\mathcal{L}_{\mathrm{repr}}$ shapes how information is
organized inside the bottleneck rather than what it reconstructs; the next
section motivates this term.

\subsection{Organizing the latent for generation}
\label{sec:method_latent_shaping}
\begin{table}[H]
\vspace{-1em}
\caption{\textbf{Three-step latent shaping on DA3-GIANT.}
Fusion and reconstruction repair the conditioning of the raw hierarchy; token
alignment improves transport and semantic organization but collapses relational
structure; $\mathcal{L}_{\mathrm{struct}}$ restores spatial and cross-view
structure while retaining gains.}
\label{tab:repr_diag_ablation}
\centering
\scriptsize
\setlength{\tabcolsep}{2.4pt}
\resizebox{\textwidth}{!}{%
\begin{tabular}{l ccc ccc ccc c}
\toprule
& \multicolumn{3}{c}{Diffusability (single-view)}
& \multicolumn{3}{c}{Diffusability (multi-view)}
& \multicolumn{3}{c}{Structure (single-view)}
& \multicolumn{1}{c}{Structure (cross-view)} \\
\cmidrule(lr){2-4}\cmidrule(lr){5-7}\cmidrule(lr){8-10}\cmidrule(lr){11-11}
Representation objective
& $\rho\downarrow$ & $\kappa\downarrow$ & effrank$\uparrow$
& $\rho\downarrow$ & $\kappa\downarrow$ & effrank$\uparrow$
& LNC@5$\uparrow$ & LDS$\uparrow$ & SRSS$\uparrow$
& xLNC$^{*}\uparrow$ \\
\midrule
Raw geometry (DA3-GIANT L0)
& 0.876 & $2.8{\times}10^{8}$ & 11.3
& 0.793 & $3.9{\times}10^{8}$ & 9.3
& 0.353 & 0.102 & 0.123
& 0.642 \\
Raw geometry (DA3-GIANT L3)
& 0.769 & $6.6{\times}10^{16}$ & 11.5
& 0.539 & $2.0{\times}10^{11}$ & 9.0
& 0.054 & 0.054 & 0.070
& 0.380 \\
\midrule
None
& 0.820 & \best{130} & \best{58.0}
& 0.630 & 580 & 20.0
& 0.310 & \snd{0.240} & \snd{0.300}
& \snd{0.590} \\
$\mathcal{L}_{\mathrm{tok}}$
& \snd{0.700} & \snd{170} & \snd{53.0}
& \snd{0.570} & \snd{450} & \snd{26.0}
& \snd{0.490} & 0.020 & 0.030
& 0.580 \\
$\mathcal{L}_{\mathrm{tok}}+\mathcal{L}_{\mathrm{struct}}$
& \best{0.674} & 227 & 51.3
& \best{0.564} & \best{356} & \best{45.2}
& \best{0.498} & \best{0.444} & \best{0.553}
& \best{0.591} \\
\bottomrule
\end{tabular}
}
\vspace{-1em}
\end{table}
Reconstruction determines what the bottleneck must preserve, but not how that
information should be organized. A reconstruction-only codec can therefore
recover RGB and geometry while still producing a latent that is difficult for
a flow model to learn. Table~\ref{tab:repr_diag_ablation} summarizes these
representation diagnostics; full metric definitions are provided in the
Sec.~\ref{sec:exp_latent} and supplementary material. The results make this gap
visible: although the reconstruction baseline is compact and well-conditioned,
its transport smoothness and semantic neighborhoods remain weak.

We first add position-wise representation supervision. Let
$\boldsymbol{\mu}_{v,i}$ be the posterior-mean token at location $i$ in view
$v$. We align a projected token with the co-located feature from a frozen
C-RADIO teacher~\citep{heinrich2025radio}. This token-level objective improves
transport smoothness ($\rho$) and semantic neighborhood consistency (LNC).
However, it treats each location independently. It can therefore align
individual tokens while destroying the relations among them, which sharply
reduces LDS and SRSS and slightly weakens cross-view correspondence.

We address this failure with a complementary relational objective. A frozen
DINOv2 teacher~\citep{oquab2024dinov2} provides pairwise similarities among
spatial locations, and we match those similarities directly in the raw
posterior space. This constrains the neighborhood geometry omitted by
position-wise alignment without requiring the student and teacher channel
dimensions to match. Table~\ref{tab:repr_diag_ablation} summarizes this
three-step progression.

With teacher tokens $\mathbf{c}_{v,i}$ and
$\mathbf{d}_{v,i}$ bilinearly aligned to the latent grid, the two terms are
\begin{align}
\mathcal{L}_{\mathrm{tok}}
&= \frac{1}{VN}\sum_{v,i}
   \left[1-\left\langle
   \widehat{g_\eta(\boldsymbol{\mu}_{v,i})},
   \hat{\mathbf{c}}_{v,i}\right\rangle\right],
   \label{eq:repr_token}\\
\mathcal{L}_{\mathrm{struct}}
&= \frac{1}{VN(N-1)}\sum_v\sum_{i\ne j}
   \left(
   \left\langle\hat{\boldsymbol{\mu}}_{v,i},
   \hat{\boldsymbol{\mu}}_{v,j}\right\rangle
   -\left\langle\hat{\mathbf{d}}_{v,i},
   \hat{\mathbf{d}}_{v,j}\right\rangle
   \right)^2,
   \label{eq:repr_struct}\\
\mathcal{L}_{\mathrm{repr}}
&= \lambda_{\mathrm{repa}}
   \left(\mathcal{L}_{\mathrm{tok}}
   +\lambda_\mu\mathcal{L}_{\mathrm{struct}}\right),
\qquad
\lambda_{\mathrm{repa}}=0.25,\quad \lambda_\mu=8.0.
\label{eq:repr_loss}
\end{align}
Here $g_\eta$ maps posterior tokens to the C-RADIO feature dimension and hats
denote $\ell_2$ normalization. C-RADIO supplies token-wise semantic
organization, while DINOv2 restores relational structure directly in the latent
used by the flow model. Unlike REPA~\citep{yu2025repa}, which supervises an
intermediate denoiser representation, both terms shape the codec latent before
generative training. The teachers and projector are discarded after codec
training.

After training, we freeze the codec, use the posterior mean
$\mathbf{z}_v:=\boldsymbol{\mu}_v$, and standardize each channel with
training-set statistics:
\begin{equation}
    \bar{\mathbf{z}}_v =
    \frac{\mathbf{z}_v-\mathbf{m}_z}{\mathbf{s}_z+\epsilon}.
    \label{eq:latent_norm}
\end{equation}

\subsection{Conditional flow matching and generation}
\label{sec:method_flow}

\paragraph{Flow matching.}
After freezing the codec, we model only its standardized posterior means. Given
target latents $\bar{\mathbf{z}}^{\mathrm{gt}}$ and noise
$\boldsymbol{\epsilon}\sim\mathcal{N}(\mathbf{0},\mathbf{I})$, we use the
linear path $\bar{\mathbf{z}}_t=(1-t)\bar{\mathbf{z}}^{\mathrm{gt}}+
t\boldsymbol{\epsilon}$ and train a conditional transformer by
\begin{equation}
    \mathcal{L}_{\mathrm{flow}}=
    \mathbb{E}\!\left[\rho(t)
    \left\|\hat{\mathbf{u}}_\psi(\bar{\mathbf{z}}_t,t,\mathcal{C})-
    (\boldsymbol{\epsilon}-\bar{\mathbf{z}}^{\mathrm{gt}})\right\|_2^2\right].
    \label{eq:flow_loss}
\end{equation}
Following RAEv2~\citep{raev22026}, the network uses clean-latent prediction,
converted to the velocity in Eq.~\ref{eq:flow_loss}. A single transformer jointly
models all target views, allowing cross-view interaction directly in the compact
geometry-native state. Parameterization and sampling details are given in
the supplementary material.

\paragraph{Reference, camera, and text conditioning.}
\label{sec:method_cond}

Geometry encoders are set-conditioned: the feature of a reference image
encoded jointly with target views differs from that of the same image encoded
alone. Placing a full-set reference latent directly in the flow state therefore
introduces a train/inference mismatch, since the full-set reference features
available during training cannot be constructed at test time. Instead, GAE
jointly encodes the $K$ observed reference views using only the observed
references as context, and prepends the resulting clean latents
$\bar{\mathbf{z}}^{\mathrm{ref}}$ as conditioning tokens. All $V$ output
slots, including those at reference camera poses, remain noisy flow variables.
The clean reference tokens $\bar{\mathbf{z}}^{\mathrm{ref}}$ are assigned
timestep $t=0$ and attend jointly with all $V$ noisy state tokens, but are
removed before the decoder prediction head. This design avoids clamping
reference slots in the ODE, yielding a single standard Euler sampler without
a denoising-level mismatch. Camera control is provided by metric
Pl\"ucker-ray embeddings in query/key self-attention~\citep{scope2026}, while
frozen language-model features enter through cross-attention. Condition
dropout supports text-only, camera-controlled, and reference-conditioned
generation with the same weights. Precise token and ray construction is
deferred to the supplementary material.

\paragraph{Training and inference.}
\label{sec:method_unified}

Stage~1 trains the codec and RGB decoder on mixed single- and multi-view data
while keeping the geometry backbone and DPT head frozen. Stage~2 freezes the
complete codec and trains the DiT flow model on both text-to-image (T2I) and
view-conditioned generation, with variable view/reference counts and condition
dropout. At inference, Gaussian target-view latents are integrated from
$t=1$ to $0$, denormalized, and decoded into RGB and geometry through
Eq.~\ref{eq:ngd_readout}. Thus, one flow model generates a shared latent state
without a hierarchy-level cascade or a separate geometry estimator. Architecture,
optimization, and solver details appear in the supplementary material.

\section{Experiments}
\label{sec:experiments}

We evaluate the representation before testing the generator built on it. We
first analyze latent-space properties (Section~\ref{sec:exp_latent}) and codec
reconstruction (Section~\ref{sec:exp_recon}). We then evaluate
camera-controllable RGB and geometry generation with matched flow models
(Section~\ref{sec:exp_gen}), followed by ablations of the codec and
conditioning designs (Section~\ref{sec:exp_ablation}).

\paragraph{Experimental setup.}
The controlled comparison includes two pixel codecs (the single-image SD-VAE~\citep{rombach2022ldm}
and the video WAN2.1 VAE), the official semantic representation autoencoder
RAEV2~\citep{raev22026} with its frozen DINOv3-L encoder~\citep{simeoni2025dinov3},
the raw DA3-GIANT layers L0 and L3~\citep{lin2026da3}, and our codecs
\method-64 and \method-128, which compress the full DA3 hierarchy and retain its
native geometry readout. Every controlled variant is trained as a matched flow
model on RealEstate10K~\citep{zhou2018re10k} and DL3DV~\citep{ling2024dl3dv},
using the same flow-model family, camera conditioning, training budget, and
sampling protocol, and is evaluated on a shared held-out pool of 64 scenes
with nine views per scene at $252^2$, one reference view, 50 Euler steps, and
CFG $=2$. Section~\ref{sec:exp_qual_extra} instead shows a separately trained
final model, using larger-scale data, higher resolution, and more frames.
Those results are qualitative and are not part of the comparison.

\paragraph{Ranking convention.}
Blue cells mark the best and second-best controlled variants, and displayed
ties are ordered using unrounded values unless a table states otherwise. Gray
rows are external systems that are excluded from this ranking, because they
are complete pipelines rather than latents dropped into our shared flow model:
GLD~\citep{gld2026} generates a cascade of DA3 feature levels, and
Gen3R~\citep{gen3r2026} is a pretrained geometry-generation system.

\subsection{Latent-space study}
\label{sec:exp_latent}

\begin{wrapfigure}{r}{0.5\textwidth}
    \vspace{-0.5em}
    \centering
    \includegraphics[width=\linewidth]{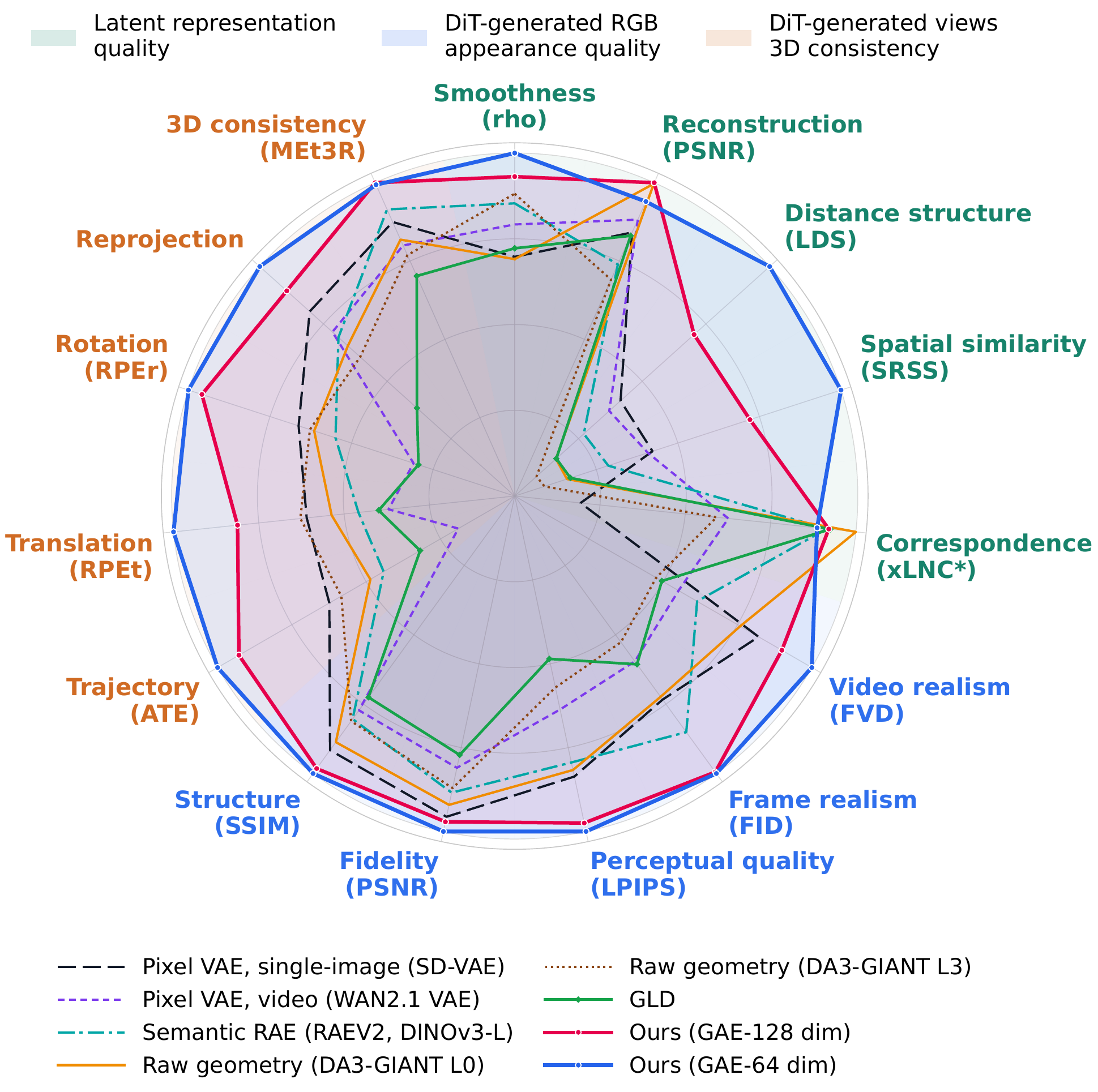}
    \vspace{-1.8em}
    \caption{\textbf{Latent representation quality and matched DiT generation.}
    Normalized overview of representation (green), RGB generation (blue), and
    3D consistency (orange). GLD uses stage-averaged representation diagnostics
    and measured reconstruction and generation results.}
    \label{fig:output_quality_radar}
    \vspace{-1.0em}
\end{wrapfigure}

We measure transport complexity with $\rho$ (rho)~\citep{zhong2026diffusing}, spectral conditioning with
$\kappa$ and effective rank, and representation structure with LNC@5~\citep{zhong2026diffusing}, LDS~\citep{raev22026,singh2025matters},
SRSS~\citep{zhong2026diffusing,singh2025matters}, and cross-view retrieval xLNC$^{*}$. Lower is better for $\rho$ and
$\kappa$; higher is better otherwise. Full definitions are in the supplementary
material.

Figure~\ref{fig:output_quality_radar} provides a compact overview of intrinsic
representation quality and matched DiT generation; Table~\ref{tab:latent_diagnostics}
reports the raw diagnostics. No baseline is strong on every axis. Pixel VAEs
are well-conditioned but weakly structured, semantic RAEv2 has strong semantic
neighborhoods but no native geometry readout, and raw DA3 features are
geometry-native but high-dimensional and poorly conditioned. Within DA3, L0
retains the strongest cross-view correspondence, whereas L3 lowers $\rho$ at
the cost of spatial and cross-view structure. Both use 3,072 channels but have
an effective rank of only about 10. By fusing the hierarchy, \method reduces
the state to 64 or 128 channels and provides the best overall balance of
transport, conditioning, and structure among the geometry-native
representations. For representation diagnostics, GLD is reported as an L0/L1
stage-average proxy because it models the two levels with a cascade.

\begin{table}[!ht]
\caption{\textbf{Latent-space diagnostics.}
\method combines smooth transport and good conditioning with strong spatial and
cross-view structure while retaining useful semantic organization.}
\label{tab:latent_diagnostics}
\centering
\scriptsize
\setlength{\tabcolsep}{3pt}
\resizebox{\textwidth}{!}{%
\begin{tabular}{l c ccc ccc ccc c}
\toprule
 & & \multicolumn{3}{c}{Diffusability (single-view)} & \multicolumn{3}{c}{Diffusability (multi-view)} & \multicolumn{3}{c}{Structure (single-view)} & \multicolumn{1}{c}{Structure (cross-view)} \\
\cmidrule(lr){3-5}\cmidrule(lr){6-8}\cmidrule(lr){9-11}\cmidrule(lr){12-12}
Latent                                                    & $C{\times}H{\times}W$      & $\rho\downarrow$ & $\kappa\downarrow$   & effrank$\uparrow$ & $\rho\downarrow$ & $\kappa\downarrow$   & effrank$\uparrow$ & LNC@5$\uparrow$ & LDS$\uparrow$ & SRSS$\uparrow$ & xLNC$^{*}\uparrow$ \\
\midrule
Pixel VAE, single-image (SD-VAE)                          & $4{\times}32{\times}32$    & 0.888            & \best{3.7}           & 3.0               & 0.765            & \best{4.1}           & 3.1               & 0.051           & 0.262         & 0.324          & 0.124 \\
Pixel VAE, video (WAN2.1 VAE)                             & $16{\times}32{\times}32$   & 0.847            & 127                  & 5.3               & 0.609            & 160                  & 4.3               & 0.084           & 0.234         & 0.313          & 0.402 \\
Semantic RAE (RAEV2)                                      & $1024{\times}16{\times}16$ & 0.711            & $1.8{\times}10^{4}$  & \best{289.3}      & 0.640            & $6.0{\times}10^{4}$  & \best{93.3}       & \best{0.932}    & 0.171         & 0.220          & 0.587 \\
Raw geometry (DA3-GIANT L0)                               & $3072{\times}18{\times}18$ & 0.876            & $2.8{\times}10^{8}$  & 11.3              & 0.793            & $3.9{\times}10^{8}$  & 9.3               & 0.353           & 0.102         & 0.123          & \best{0.642} \\
Raw geometry (DA3-GIANT L3)                               & $3072{\times}18{\times}18$ & 0.769            & $6.6{\times}10^{16}$ & 11.5              & \snd{0.539}      & $2.0{\times}10^{11}$ & 9.0               & 0.054           & 0.054         & 0.070          & 0.380 \\
\rowcolor{gray!25} GLD (L0/L1 stage avg.) & --                         & 0.854            & $9.5{\times}10^{7}$  & 34.8              & 0.742            & $1.7{\times}10^{8}$  & 22.7              & 0.274           & 0.103         & 0.131          & 0.594 \\
\midrule
Ours (\method-128)                                            & $128{\times}18{\times}18$  & \snd{0.674}      & 227                  & \snd{51.3}        & 0.564            & 356                  & \snd{45.2}        & \snd{0.498}     & \snd{0.444}   & \snd{0.553}    & \snd{0.591} \\
Ours (\method-64)                                             & $64{\times}18{\times}18$   & \best{0.634}     & \snd{53.1}           & 37.0              & \best{0.519}     & \snd{117}            & 33.2              & 0.488           & \best{0.631}  & \best{0.767}   & 0.569 \\
\bottomrule
\end{tabular}%
}
\end{table}

\subsection{Codec reconstruction quality}
\label{sec:exp_recon}

\subsubsection{RGB reconstruction}
\begin{wraptable}[10]{r}{0.50\textwidth}
\vspace{-2em}
\centering
\caption{\textbf{RGB reconstruction.} Single-view uses ImageNet; multi-view
averages five video datasets.}
\label{tab:recon}
\footnotesize
\setlength{\tabcolsep}{2.5pt}
\resizebox{\linewidth}{!}{%
\begin{tabular}{l ccc ccc}
\toprule
& \multicolumn{3}{c}{Single-view} & \multicolumn{3}{c}{Multi-view} \\
\cmidrule(lr){2-4}\cmidrule(lr){5-7}
Latent & PSNR$\uparrow$ & LPIPS$\downarrow$ & rFID$\downarrow$ & PSNR$\uparrow$ & LPIPS$\downarrow$ & rFVD$\downarrow$ \\
\midrule
Pixel VAE, single-image (SD-VAE)     & 24.44& 0.141 & 26.9 & 28.97 & 0.112 &40.5\\
Pixel VAE, video (WAN2.1 VAE)     & 27.49 &0.125 & 21.8   &28.12 & 0.156 & 50.6 \\
Semantic RAE (RAEV2)     & 22.08 & 0.134 & 8.8  & 24.73 & 0.094 & 32.1 \\
Raw geometry (DA3-GIANT L0)   & \snd{28.64} & \snd{0.038} & \snd{5.5} & \snd{34.70} & \best{0.014} & \snd{4.4} \\
Raw geometry (DA3-GIANT L3)     & 19.90 & 0.219 & 25.3 & 23.98 & 0.122 & 76.2 \\
\midrule
Ours (\method-128)  & \best{28.76} & \best{0.036} & \best{5.4} & \best{34.78} & \best{0.014} & \best{4.3} \\
Ours (\method-64)   & 27.30 & 0.061 & 7.1 & 32.37 & \snd{0.040} & 9.1 \\
\bottomrule
\end{tabular}%
}
\vspace{-0.8em}
\end{wraptable}

We report PSNR and LPIPS for paired fidelity and rFID/rFVD for the
distributional gap between reconstructed and real images/videos. Higher PSNR
and lower LPIPS, rFID, and rFVD are better.

A compact latent risks discarding the fine-grained spatial detail retained by
raw geometry features. We therefore evaluate codec reconstruction
independently of downstream generation. Table~\ref{tab:recon} measures RGB
fidelity. A separately trained RGB head on raw DA3 L0 reconstructs appearance
well, but the same $3072$-channel feature is ill-conditioned and nearly
rank-degenerate (Table~\ref{tab:latent_diagnostics}). Raw L3 loses
substantially more reconstructable appearance information. \method-128 matches or
slightly exceeds raw L0 across PSNR, LPIPS, rFID, and rFVD while using 24
times fewer channels.

\subsubsection{Geometry reconstruction}
\label{geo_recons}
Table~\ref{tab:geom_recon} evaluates geometry after encoding and decoding real
frames. AbsRel and $\delta_1$ measure relative depth error and threshold
accuracy; Chamfer and PMap measure point-set and point-map discrepancy; ATE
and RPEr measure trajectory and relative-rotation error. Only $\delta_1$ is
higher-is-better. The raw-L0 baseline (DA3-GIANT L0)
propagates L0 through the frozen DA3 backbone to recover the remaining
hierarchy before the DPT head. Both \method variants improve depth, point-cloud,
point-map, and absolute trajectory accuracy over this baseline on both datasets,
while relative-rotation errors remain comparable. Thus, the compact
codec preserves geometry-relevant information in addition to the RGB fidelity
measured above. Compared with the external systems, \method gives lower camera
and point-cloud errors than Gen3R~\citep{gen3r2026} and GLD~\citep{gld2026};
Gen3R is slightly better on the depth metrics but uses a separate geometry
latent. Pixel codecs and RAEV2 are
omitted because they have no native geometry readout.

\begin{table}[!ht]
\caption{\textbf{Geometry reconstruction from encoded latents.}
Depth and point maps are compared with Pi3~\citep{wang2026pi3} on the real target frames; camera
poses are compared with the input real trajectory. Displayed ties between \method-64 and \method-128 are ordered using
unrounded values.}
\label{tab:geom_recon}
\centering
\scriptsize
\setlength{\tabcolsep}{3pt}
\resizebox{\textwidth}{!}{%
\begin{tabular}{l cccccc cccccc}
\toprule
 & \multicolumn{6}{c}{RealEstate10K (Pi3 ref)} & \multicolumn{6}{c}{DL3DV (Pi3 ref)} \\
\cmidrule(lr){2-7}\cmidrule(lr){8-13}
Method & AbsRel$\downarrow$ & $\delta_1\uparrow$ & Cham.$\downarrow$ & PMap$\downarrow$ & ATE$\downarrow$ & RPEr$^{\circ}\downarrow$
 & AbsRel$\downarrow$ & $\delta_1\uparrow$ & Cham.$\downarrow$ & PMap$\downarrow$ & ATE$\downarrow$ & RPEr$^{\circ}\downarrow$ \\
\midrule
Pi3 on real (ref)                          & ---          & ---          & ---          & ---          & 0.009        & 0.17        & ---          & ---          & ---          & ---          & 0.011        & 0.26 \\
\midrule
\rowcolor{gray!25} Gen3R & 0.085        & 0.906        & 0.412        & 0.803        & 0.038        & 0.63        & 0.114        & 0.863        & 0.237        & 0.541        & 0.022        & 0.72 \\
\rowcolor{gray!25} GLD     & 0.109        & 0.873        & 0.516        & 0.938        & 0.021        & 0.47        & 0.135        & 0.839        & 0.208        & 0.543        & 0.016        & 0.45 \\
Raw geometry (DA3-GIANT L0) & 0.091    & 0.900 & 0.526 & 0.923 & 0.008 & 0.22 &0.127    & 0.858 & 0.182    & 0.482  &   0.009 & \snd{0.26} \\
Ours (\method-128)                             & \snd{0.090}  & \snd{0.905}  & \snd{0.400}  & \snd{0.700}  & \best{0.006} & \best{0.19} & \snd{0.125}  & \best{0.862} & \best{0.145} & \best{0.433} & \best{0.008} & \best{0.25} \\
Ours (\method-64)                              & \best{0.090} & \best{0.905} & \best{0.386} & \best{0.691} & \snd{0.007}  & \snd{0.20}  & \best{0.125} & \snd{0.861}  & \snd{0.145}  & \snd{0.433}  & \snd{0.008}  & 0.27 \\
\bottomrule
\end{tabular}%
}
\end{table}

\subsection{Flow-matching generation quality}
\label{sec:exp_gen}

We compare all latent families under the same flow-matching backbone, training budget, and camera-conditioning mechanism. This controlled setup isolates the impact of the generated latent representation on both appearance quality and 3D consistency. 

\subsubsection{RGB generation}
\begin{figure}[!ht]
\centering
\includegraphics[width=\linewidth]{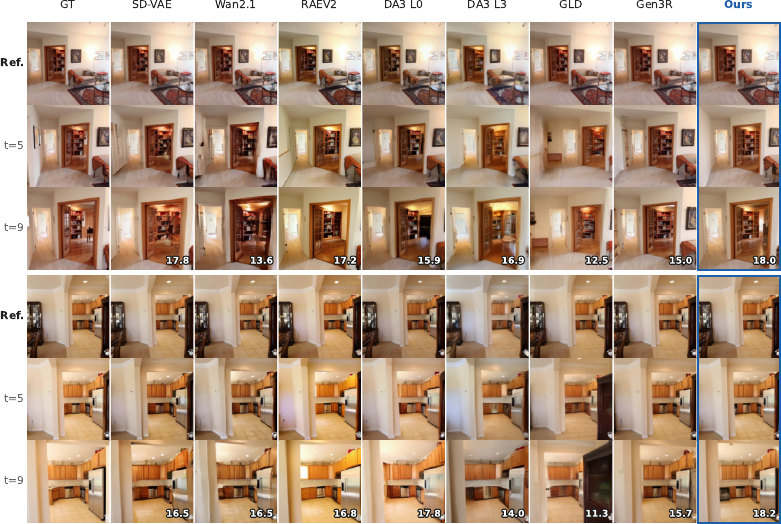}
\vspace{-2em}
\caption{\textbf{Qualitative RGB comparison on RealEstate10K.}
Columns are methods and rows show one reference and two target views for each
scene under identical camera conditioning. Insets report per-scene PSNR over
the two displayed target views; Ours denotes the $C{=}64$ model.}
\label{fig:qual_rgb}
\end{figure}

Table~\ref{tab:latent_ablation} evaluates distributional quality with FID/FVD
and paired fidelity with LPIPS, PSNR, and SSIM. \method-64 gives the strongest
overall results on both datasets, reducing FVD by $12.7\%$ on RealEstate10K and
$23.1\%$ on DL3DV relative to the best non-\method controlled latent. It also
leads the paired metrics, while \method-128 gives the best RealEstate10K FID. Figure~\ref{fig:qual_rgb} provides the corresponding RGB comparison on two RealEstate10K scenes. Under the same reference image and camera trajectory, \method-64 better preserves object identity and fine structure as the viewpoint moves away from the reference, consistent with the metrics in Table~\ref{tab:latent_ablation}.

\begin{table}[!ht]
\caption{\textbf{Camera-conditioned appearance generation.}
All controlled latent variants use the matched protocol described above.}
\label{tab:latent_ablation}
\centering
\scriptsize
\setlength{\tabcolsep}{3pt}
\resizebox{\textwidth}{!}{%
\begin{tabular}{l ccccc ccccc}
\toprule
 & \multicolumn{5}{c}{RealEstate10K} & \multicolumn{5}{c}{DL3DV} \\
\cmidrule(lr){2-6}\cmidrule(lr){7-11}
Latent & FVD$\downarrow$ & FID$\downarrow$ & LPIPS$\downarrow$ & PSNR$\uparrow$ & SSIM$\uparrow$
 & FVD$\downarrow$ & FID$\downarrow$ & LPIPS$\downarrow$ & PSNR$\uparrow$ & SSIM$\uparrow$ \\
\midrule
Pixel VAE, single-image (SD-VAE)           & 258.6        & 36.4        & 0.171        & 19.32        & 0.666        & 373.2        & 57.1        & 0.232        & 17.05        & 0.495 \\
Pixel VAE, video (WAN2.1 VAE)              & 362.9        & 41.0        & 0.219        & 16.52        & 0.575        & 596.7        & 77.3        & 0.313        & 14.31        & 0.407 \\
Semantic RAE (RAEV2)                       & 379.4        & 31.4        & 0.191        & 17.60        & 0.585        & 453.4        & 49.2        & 0.231        & 16.01        & 0.434 \\
Raw geometry (DA3-GIANT L0)                & 298.6        & 35.5        & 0.177        & 18.31        & 0.643        & 376.5        & 60.2        & 0.235        & 16.68        & 0.481 \\
Raw geometry (DA3-GIANT L3)                & 488.9        & 50.0        & 0.273        & 16.82        & 0.579        & 584.8        & 80.6        & 0.314        & 16.22        & 0.447 \\
\midrule
Ours (\method-128)                             & \snd{233.4}  & \best{25.6} & \snd{0.146}  & \snd{19.54}  & \snd{0.701}  & \snd{345.2}  & \snd{44.2}  & \snd{0.200}  & \snd{17.39}  & \snd{0.541} \\
Ours (\method-64)                              & \best{225.7} & \snd{27.1}  & \best{0.143} & \best{20.02} & \best{0.711} & \best{287.0} & \best{41.3} & \best{0.194} & \best{18.00} & \best{0.553} \\
\midrule
\rowcolor{gray!25} GLD     & 446.3        & 45.4        & 0.337        & 14.85        & 0.517        & 591.8        & 67.1        & 0.355        & 14.43        & 0.400 \\
\rowcolor{gray!25} Gen3R & 266.4        & 26.9        & 0.195        & 17.73        & 0.623        & 365.4        & 42.5        & 0.247        & 16.16        & 0.479 \\
\bottomrule
\end{tabular}%
}
\end{table}

\subsubsection{3D consistency}
\begin{figure}[!ht]
\centering
\includegraphics[width=\linewidth]{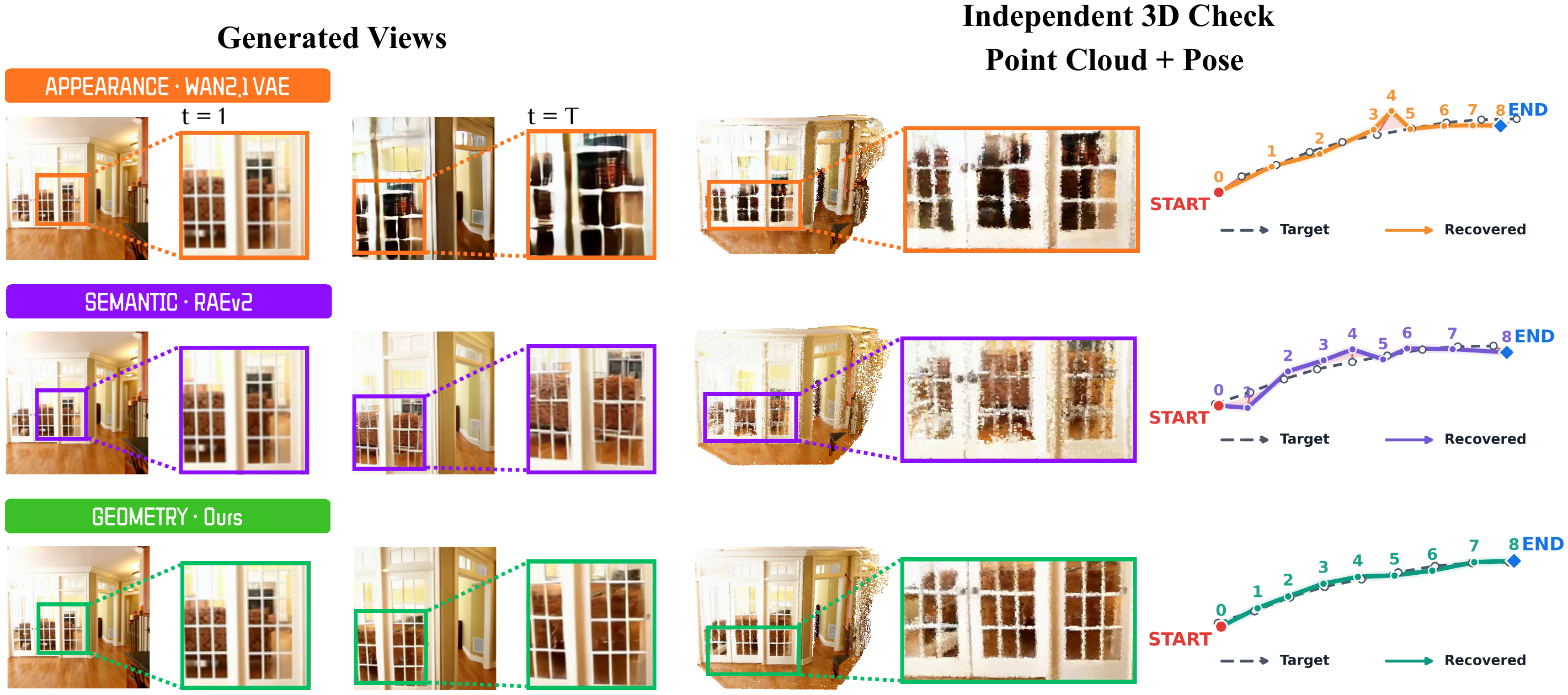}
\vspace{-2em}
\caption{\textbf{Independent 3D check across latent families.}
The appearance-native Wan2.1 VAE, semantic-native RAEv2, and geometry-native
GAE share the same reference image and camera conditioning. Generated views
and detail crops are shown alongside point clouds and camera trajectories
reconstructed by VGGT. The recovered trajectories are compared with those
recovered from the corresponding real videos, highlighting cross-view visual
and camera-motion consistency.}
\label{fig:qual_rgb_dl3dv}
\end{figure}

Table~\ref{tab:consistency} evaluates camera and multi-view consistency using
VGGT~\citep{wang2025vggt}, MEt3R~\citep{asim2025met3r}, and a
DA3-GIANT~\citep{lin2026da3} cross-check. We apply the same estimator to the
generated and corresponding real videos. ATE measures the discrepancy between
their recovered trajectories after Sim(3) alignment, while RPEt and RPEr measure
relative translation and rotation errors, respectively, with RPEr in degrees.
For reprojection error~\citep{gld2026}, we use the recovered world points,
cameras, and intrinsics to measure the normalized 2D round-trip pixel error
between high-confidence points across
views, retaining the top $50\%$ confidence pixels. MEt3R~\citep{asim2025met3r} directly measures
symmetric feature disagreement between reconstructed view pairs with
visibility and occlusion masking. 
Across both datasets, the two \method variants achieve the two best controlled
VGGT results. \method-64 reduces ATE by $52.8\%$ on RealEstate10K and
$23.3\%$ on DL3DV, while \method-64 and \method-128 achieve the best controlled
MEt3R on RealEstate10K ($0.1208$) and DL3DV ($0.1347$), respectively. Together,
these results show that the appearance gains preserve camera-motion and
multi-view consistency. Compared with the external systems, both \method
variants have lower ATE on both datasets; Gen3R achieves lower MEt3R on
RealEstate10K, whereas both \method variants achieve lower MEt3R on DL3DV.
Figure~\ref{fig:qual_rgb_dl3dv} further shows that \method yields a more
coherent 3D scene and camera motion more consistent with the corresponding
real video.
\begin{table}[!ht]
\caption{\textbf{Generated-view 3D consistency}.
VGGT is independent of all latent backbones; DA3-GIANT is reported as a
backbone-overlap cross-check. Camera errors compare trajectories recovered from
generated and corresponding real videos using the same estimator.}
\label{tab:consistency}
\centering
\scriptsize
\setlength{\tabcolsep}{2.8pt}
\resizebox{\textwidth}{!}{%
\begin{tabular}{l cccc cccc c}
\toprule
 & \multicolumn{4}{c}{VGGT (independent)} & \multicolumn{4}{c}{DA3-GIANT (cross-check)} & \\
\cmidrule(lr){2-5}\cmidrule(lr){6-9}
Latent                                     & ATE$\downarrow$ & RPEt$\downarrow$ & RPEr$^{\circ}\downarrow$ & Reproj.$\downarrow$ & ATE$\downarrow$ & RPEt$\downarrow$ & RPEr$^{\circ}\downarrow$ & Reproj.$\downarrow$ & MEt3R$\downarrow$ \\
\midrule
\multicolumn{10}{l}{\emph{RealEstate10K}} \\
Pixel VAE, single-image (SD-VAE)           & 0.0072          & 0.0182           & 0.648                    & 0.0093              & 0.0046          & 0.0191           & 0.338                    & 0.0018              & 0.1385 \\
Pixel VAE, video (WAN2.1 VAE)              & 0.0225          & 0.0252           & 0.746                    & 0.0105              & 0.0135          & 0.0176           & 0.751                    & 0.0019              & 0.1441 \\
Semantic RAE (RAEV2)                       & 0.0085          & 0.0233           & 0.420                    & 0.0104              & 0.0066          & 0.0219           & 0.507                    & 0.0019              & 0.1263 \\
Raw geometry (DA3-GIANT L0)                & 0.0085          & 0.0202           & 0.435                    & 0.0096              & 0.0047          & 0.0167           & 0.358                    & 0.0020              & 0.1420 \\
Raw geometry (DA3-GIANT L3)                & 0.0112          & 0.0211           & 0.516                    & 0.0124              & 0.0036          & 0.0148           & 0.355                    & 0.0019              & 0.1575 \\
\midrule
Ours (\method-128)                             & \snd{0.0041}    & \snd{0.0173}     & \best{0.332}             & \snd{0.0075}        & \best{0.0026}   & \snd{0.0143}     & \snd{0.304}              & \snd{0.0017}        & \snd{0.1213} \\
Ours (\method-64)                              & \best{0.0034}   & \best{0.0060}    & \snd{0.380}              & \best{0.0068}       & \best{0.0026}   & \best{0.0141}    & \best{0.225}             & \best{0.0013}       & \best{0.1208} \\
\midrule
\rowcolor{gray!25} GLD     & 0.0166          & 0.0270           & 0.837                    & 0.0133              & 0.0098          & 0.0237           & 0.878                    & 0.0035              & 0.1572 \\
\rowcolor{gray!25} Gen3R & 0.0079          & 0.0178           & 0.512                    & 0.0061              & 0.0060          & 0.0181           & 0.361                    & 0.0013              & 0.1145 \\
\midrule
\multicolumn{10}{l}{\emph{DL3DV}} \\
Pixel VAE, single-image (SD-VAE)           & 0.0073          & 0.0098           & 0.432                    & 0.0108              & 0.0055          & 0.0076           & 0.358                    & \snd{0.0024}        & 0.1542 \\
Pixel VAE, video (WAN2.1 VAE)              & 0.0243          & 0.0288           & 1.419                    & 0.0258              & 0.0206          & 0.0251           & 1.016                    & \best{0.0019}       & 0.1777 \\
Semantic RAE (RAEV2)                       & 0.0104          & 0.0143           & 0.638                    & 0.0130              & 0.0088          & 0.0113           & 0.585                    & 0.0031              & 0.1549 \\
Raw geometry (DA3-GIANT L0)                & 0.0110          & 0.0136           & 0.634                    & 0.0129              & 0.0079          & 0.0102           & 0.482                    & 0.0042              & 0.1721 \\
Raw geometry (DA3-GIANT L3)                & 0.0097          & 0.0120           & 0.603                    & 0.0151              & 0.0052          & 0.0073           & 0.380                    & 0.0039              & 0.1754 \\
\midrule
Ours (\method-128)                             & \snd{0.0056}    & \snd{0.0072}     & \snd{0.294}              & \snd{0.0080}        & \snd{0.0043}    & \snd{0.0053}     & \snd{0.260}              & 0.0029              & \best{0.1347} \\
Ours (\method-64)                              & \best{0.0056}   & \best{0.0066}    & \best{0.293}             & \best{0.0077}       & \best{0.0038}   & \best{0.0053}    & \best{0.254}             & 0.0028              & \snd{0.1370} \\
\midrule
\rowcolor{gray!25} GLD     & 0.0125          & 0.0159           & 1.216                    & 0.0245              & 0.0107          & 0.0146           & 1.020                    & 0.0089              & 0.2127 \\
\rowcolor{gray!25} Gen3R & 0.0096          & 0.0127           & 0.562                    & 0.0088              & 0.0102          & 0.0128           & 0.510                    & 0.0050              & 0.1456 \\
\bottomrule
\end{tabular}
}
\end{table}

\subsubsection{Geometry generation}
Table~\ref{tab:geom_gen} evaluates geometry decoded from sampled latents using
the metrics defined in Section~\ref{geo_recons}. We directly decode depth,
camera rays, and point maps from each latent. Depth and point maps are
compared with Pi3~\citep{wang2026pi3} applied to the real target frames, while camera trajectories
are compared with the input dataset poses. This common reference avoids evaluating
each method against geometry estimated from its own generated images.

\begin{figure}[!ht]
\centering
\includegraphics[width=\linewidth]{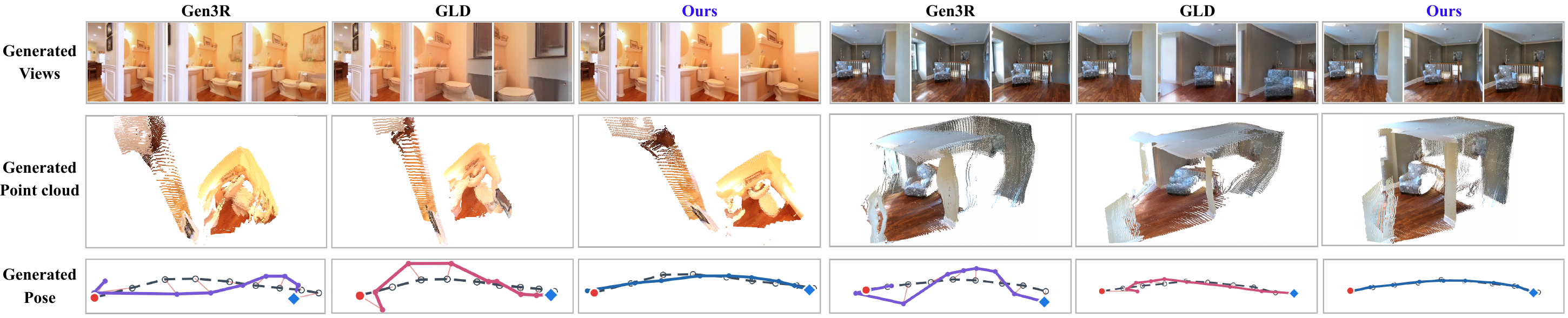}
\vspace{-2em}
\caption{\textbf{Multi-frame RGB, geometry, and camera-pose comparison across
RealEstate10K scenes.}
Gen3R, GLD, and Ours receive the same reference image and prescribed camera
trajectory. For each scene, early, middle, and final generated views are
paired with the generated point cloud and recovered pose. The multi-scene
comparison exposes cross-frame structure, depth-layer distortions, and
trajectory drift that are less apparent in a single final view.}
\label{fig:joint_rgb_geometry}
\end{figure}

On DL3DV, \method-64 leads all six geometry metrics among the controlled
geometry-native methods. On RealEstate10K, it gives the best depth accuracy
and pose estimates, while \method-128 gives lower Chamfer and point-map errors.
Because the depth and point references come from Pi3 and the pose reference
comes from the dataset cameras, these gains do not rely on DA3-GIANT as the
evaluator. Figure~\ref{fig:joint_rgb_geometry} focuses on comparisons with external
baselines. By showing multiple frames, generated point clouds, and recovered
camera poses for the same scenes, it makes cross-frame drift and geometric
distortions readily visible. GAE closely follows the prescribed camera
trajectory and preserves the overall scene structure, while Gen3R and GLD
exhibit larger pose deviations and more severe point-cloud distortions.

\begin{table}[!ht]
\caption{\textbf{Geometry generation from sampled latents.}
Depth and point maps are compared with Pi3 on the real target frames; camera
poses are compared with the real trajectory.}
\label{tab:geom_gen}
\centering
\scriptsize
\setlength{\tabcolsep}{3pt}
\resizebox{\textwidth}{!}{%
\begin{tabular}{l cccccc cccccc}
\toprule
 & \multicolumn{6}{c}{RealEstate10K (Pi3 ref)} & \multicolumn{6}{c}{DL3DV (Pi3 ref)} \\
\cmidrule(lr){2-7}\cmidrule(lr){8-13}
Method & AbsRel$\downarrow$ & $\delta_1\uparrow$ & Cham.$\downarrow$ & PMap$\downarrow$ & ATE$\downarrow$ & RPEr$^{\circ}\downarrow$
 & AbsRel$\downarrow$ & $\delta_1\uparrow$ & Cham.$\downarrow$ & PMap$\downarrow$ & ATE$\downarrow$ & RPEr$^{\circ}\downarrow$ \\
\midrule
Pi3 on real (ref)                          & ---          & ---          & ---          & ---          & 0.009        & 0.17        & ---          & ---          & ---          & ---          & 0.011        & 0.26 \\
\midrule
\rowcolor{gray!25} Gen3R & 0.169        & 0.792        & 0.712        & 1.325        & 0.035        & 0.73        & 0.221        & 0.687        & 0.570        & 1.179        & 0.032        & 0.87 \\
\rowcolor{gray!25} GLD    & 0.259        & 0.696        & 0.944        & 1.742        & 0.037        & 1.07        & 0.318        & 0.598        & 0.487        & 1.152        & 0.027        & 1.20 \\
Raw geometry (DA3-GIANT L0)                & 0.157        & 0.822        & 0.540        & 1.058        & 0.019        & \snd{0.35}  & 0.245        & 0.715        & 0.330        & \snd{0.815}  & 0.019        & 0.48 \\
Ours (\method-128)                             & \snd{0.139}  & \snd{0.849}  & \best{0.432} & \best{0.813} & \snd{0.013}  & 0.37        & \snd{0.225}  & \snd{0.750}  & \snd{0.325}  & 0.832        & \snd{0.016}  & \snd{0.40} \\
Ours (\method-64)                              & \best{0.134} & \best{0.857} & \snd{0.501}  & \snd{0.924}  & \best{0.010} & \best{0.33} & \best{0.204} & \best{0.765} & \best{0.262} & \best{0.698} & \best{0.014} & \best{0.36} \\
\bottomrule
\end{tabular}%
}
\end{table}

\subsubsection{Additional qualitative capabilities}
\label{sec:exp_qual_extra}
\begin{figure}[!ht]
\centering
\includegraphics[width=\linewidth]{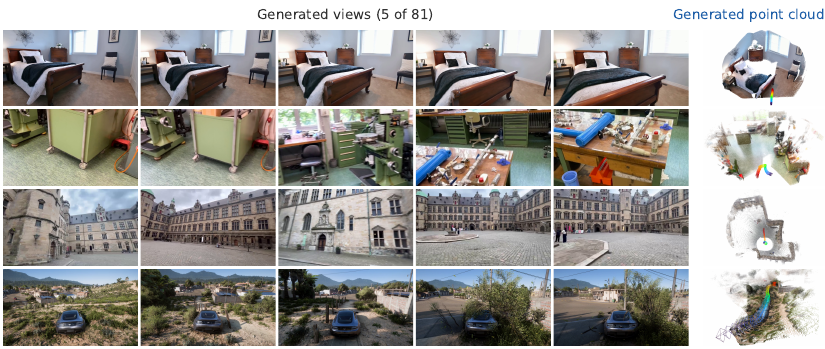}
\vspace{-2em}
\caption{\textbf{Long-rollout showcase.}
Five views from each 81-view rollout and the point cloud decoded from the same
latents.}
\label{fig:showcase_geoft}
\end{figure}

\paragraph{Long-rollout showcase.}
The controlled comparisons above use nine views. Figure~\ref{fig:showcase_geoft} instead shows an 81-view rollout at $672{\times}378$, sampled from one reference view with 50 Euler steps and CFG $=2$. We decode each sampled latent once into RGB and geometry. The point cloud simply concatenates the 81 per-view depth maps after unprojection with cameras recovered from the decoded rays; it uses no cross-view fusion, external reconstruction, or test-time optimization. Detailed training settings and dataset information are provided in the supplementary material.

\begin{figure}[!ht]
\centering
\includegraphics[width=\linewidth]{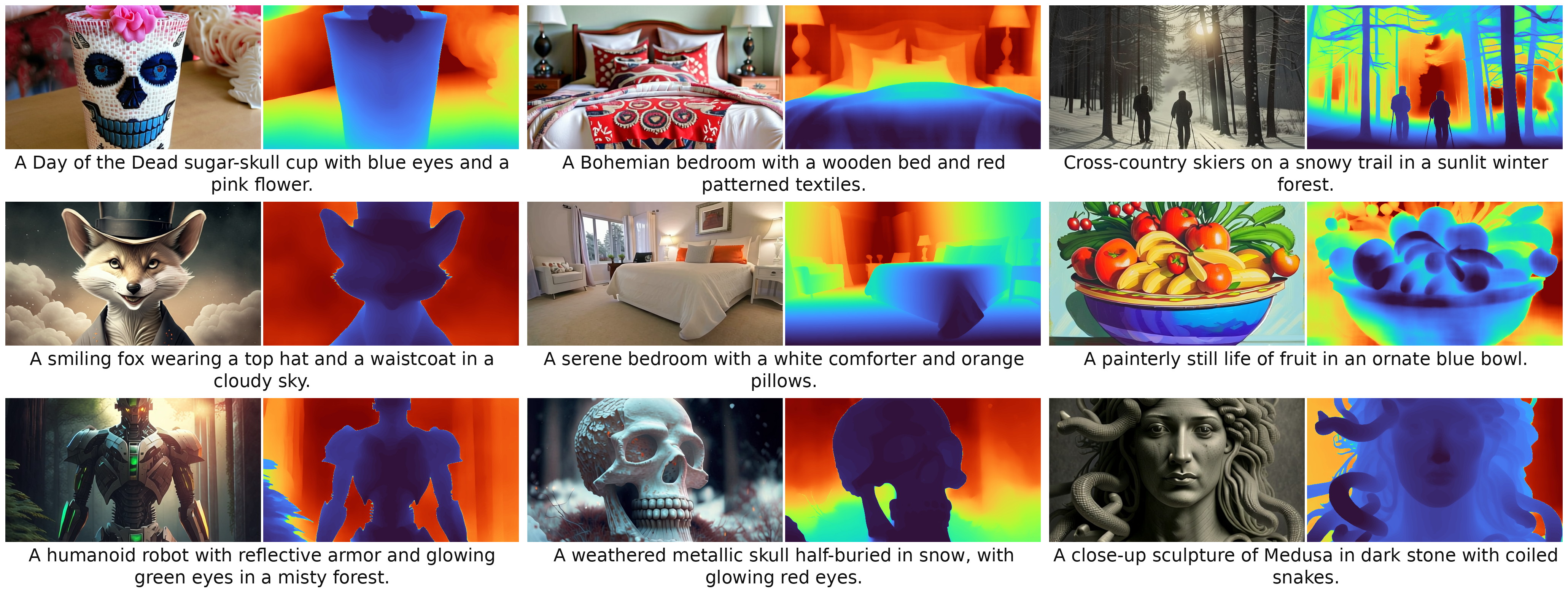}
\vspace{-2em}
\caption{\textbf{Text-conditioned RGB and geometry generation.}
Each pair shows an RGB sample and depth decoded from the same latent. These
selected examples demonstrate capability rather than benchmark text-to-image
quality.}
\label{fig:t2i_samples}
\end{figure}

\paragraph{Text-conditioned generation.}
The same flow checkpoint used for the long-rollout showcase also supports
direct text-conditioned RGB and geometry generation.
Figure~\ref{fig:t2i_samples} pairs each generated image with depth decoded
from the same latent for a compact set of diverse prompts. We treat these
samples as a qualitative capability check rather than a standalone
text-to-image benchmark.

\subsection{Ablation studies}
\label{sec:exp_ablation}

We ablate representation shaping, codec backbone, reference conditioning, pose
parameterization, and text-to-image co-training. Within each block, we vary
only the component under study while keeping the flow architecture and all
remaining training settings fixed. The representation diagnostics in
Table~\ref{tab:repr_diag_ablation} show that token-level alignment improves
latent transport but compromises pairwise structure; adding
$\mathcal{L}_{\mathrm{struct}}$ recovers this structure, allowing the complete
$\mathcal{L}_{\mathrm{repr}}$ to balance transportability, semantic organization,
and relational consistency.

\begin{wraptable}[18]{r}{0.50\textwidth}
  \vspace{-1.0em}
  \centering
  \caption{\textbf{Codec and conditioning ablations on RealEstate10K.}
  Diag. denotes ATE, point-cloud gap, scale-sensitive SE(3) ATE, or LPIPS,
  depending on the block. Except for PSNR, lower is better.}
  \label{tab:gen_ablation}
  \tiny
  \setlength{\tabcolsep}{2pt}
  \resizebox{\linewidth}{!}{%
    \begin{tabular}{lcccc}
      \toprule
      Setting
      & FVD $\downarrow$
      & PSNR $\uparrow$
      & MEt3R $\downarrow$
      & Diag. $\downarrow$ \\
      \midrule
      \multicolumn{5}{l}{%
        \emph{Codec design (Diag. $=$ VGGT ATE)}} \\
      LARGE-128, with $\mathcal{L}_{\mathrm{repr}}$
      & 270.5
      & 19.12
      & \best{0.1192}
      & 0.0048 \\
      GIANT-128, no $\mathcal{L}_{\mathrm{repr}}$
      & 266.4
      & 19.19
      & 0.1226
      & 0.0044 \\
      GIANT-128, with $\mathcal{L}_{\mathrm{tok}}$
      & 241.9
      & 19.40
      & 0.1212
      & 0.0042 \\
      GIANT-128, with
$\mathcal{L}_{\mathrm{tok}}+\mathcal{L}_{\mathrm{struct}}$
      ($\mathcal{L}_{\mathrm{repr}}$)
      & \best{233.4}
      & \best{19.54}
      & 0.1213
      & \best{0.0041} \\
      \midrule
      \multicolumn{5}{l}{%
        \emph{Reference conditioning (Diag. $=$ Ref--tgt PC gap)}} \\
      Reference latent in ODE state
      & 257.1
      & 18.91
      & 0.1217
      & 0.1016 \\
      Clean conditioning token
      & \best{225.7}
      & \best{20.02}
      & \best{0.1208}
      & \best{0.0318} \\
      \midrule
      \multicolumn{5}{l}{%
        \emph{Pose scale (Diag. $=$ SE(3) ATE, no Sim(3))}} \\
      Per-scene normalized pose
      & 350.7
      & 16.28
      & 0.1290
      & 0.0346 \\
      Metric pose
      & \best{225.7}
      & \best{20.02}
      & \best{0.1208}
      & \best{0.0093} \\
      \midrule
      \multicolumn{5}{l}{%
        \emph{Text-to-image co-training (Diag. $=$ LPIPS)}} \\
      Without T2I co-training
      & 472.9
      & 16.77
      & 0.1262
      & 0.199 \\
      With T2I co-training
      & \best{225.7}
      & \best{20.02}
      & \best{0.1208}
      & \best{0.143} \\
      \bottomrule
    \end{tabular}%
  }
  \vspace{-0.8em}
\end{wraptable}

Table~\ref{tab:gen_ablation} confirms that these design choices translate into
downstream generation gains. On DA3-GIANT, $\mathcal{L}_{\mathrm{tok}}$
improves FVD, PSNR, and camera-motion consistency; adding
$\mathcal{L}_{\mathrm{struct}}$ further improves
FVD and PSNR and camera accuracy. At the same latent width and with the complete objective,
DA3-GIANT outperforms DA3-LARGE in appearance quality and camera accuracy,
whereas DA3-LARGE retains a small MEt3R advantage. Reference conditioning is
also important: supplying the reference as a clean conditioning token, rather
than evolving it as part of the ODE state, improves every metric and reduces
the reference--target point-cloud gap by more than threefold. Preserving metric
pose scale similarly yields large gains in appearance, cross-view consistency,
and scale-sensitive trajectory accuracy. Finally, text-to-image co-training
markedly improves all reported metrics, including perceptual similarity.
Together, these results show that representation shaping, backbone capacity,
clean reference conditioning, metric pose control, and text-to-image
co-training provide complementary gains.


\section{Conclusion}
We presented \method, a compact reparameterization of multi-level geometry
features that is jointly decodable to RGB and geometry and organized for flow
modeling. Under a controlled generator and training protocol, this latent
improves reconstruction, visual generation, 3D consistency, and camera control
over pixel, semantic, and raw geometry representations. Additional qualitative
demonstrations illustrate longer, higher-resolution generation with jointly
decoded RGB and geometry. Together, these results show that
geometry can define the state in which generation occurs, providing a shared
interface between perception and generation.

\bibliography{iclr2027_conference}
\bibliographystyle{iclr2027_conference}

\appendix

\section{Implementation and evaluation details}
\label{app:method_details}

\subsection{Flow-matching parameterization}
\label{app:flow_matching}

After freezing the codec, we train only on standardized posterior means. Let
$\bar{\mathbf{z}}^{\mathrm{gt}}_{1:V}$ denote the target latent and
$\boldsymbol{\epsilon}\sim\mathcal{N}(\mathbf{0},\mathbf{I})$ Gaussian noise,
with the initial sampling state $\bar{\mathbf{z}}^{(1)}=\boldsymbol{\epsilon}$.
We interpolate between them using
\begin{equation}
    \bar{\mathbf{z}}_t=(1-t)\bar{\mathbf{z}}^{\mathrm{gt}}
    +t\boldsymbol{\epsilon},\qquad t\in[0,1],
    \label{eq:linear_path}
\end{equation}
whose target velocity is
$\mathbf{u}_t=\boldsymbol{\epsilon}-\bar{\mathbf{z}}^{\mathrm{gt}}$.
Following RAEv2~\citep{raev22026}, the network predicts the clean latent,
\begin{equation}
    \hat{\bar{\mathbf{z}}}^{\mathrm{gt}}
    =\mathbf{f}_{\psi}(\bar{\mathbf{z}}_t,t,\mathcal{C}),
\end{equation}
which is converted to velocity by
\begin{equation}
    \hat{\mathbf{u}}_t=
    \frac{\bar{\mathbf{z}}_t-\hat{\bar{\mathbf{z}}}^{\mathrm{gt}}}
         {\max(t,t_{\epsilon})},\qquad t_{\epsilon}=0.05.
    \label{eq:xpred_to_v}
\end{equation}
We optimize the flow objective
\begin{equation}
    \mathcal{L}_{\mathrm{flow}}=
    \mathbb{E}\!\left[\rho(t)
    \left\|\hat{\mathbf{u}}_\psi(\bar{\mathbf{z}}_t,t,\mathcal{C})-
    (\boldsymbol{\epsilon}-\bar{\mathbf{z}}^{\mathrm{gt}})\right\|_2^2\right],
    \label{eq:app_flow_loss}
\end{equation}
with an optional time weight $\rho(t)$.
The timestep is shared across all views in a clip. We use a DDT-style transformer~\citep{raev22026}
with $1\times1$ latent tokens and cross-view self-attention; temporal models also
use frame-axis rotary embeddings. The controlled models use the same
encoder--decoder transformer recipe: 28 encoder blocks of width 768 and six
decoder blocks of width 2048. Only the latent input/output projections change
when the codec width changes from 128 to 64 channels.

\subsection{Inference-aligned multimodal conditioning}
\label{app:multimodal_conditioning}

The conditioning design follows a single principle: every signal is constructed
from information that is genuinely available at inference time and is kept
separate from the variables evolved by the flow. In particular, only the $V$
requested output latents form the noisy ODE state. Observed images are converted
once into clean geometry-native evidence tokens, camera parameters specify a
spatial field over both observed and requested views, and text supplies global
semantic context. We denote these signals by
$\mathcal{C}=\{\bar{\mathbf{z}}^{\mathrm{ref}},\mathbf{r},
\mathbf{e}_{\mathrm{text}}\}$. They are injected through complementary
attention pathways and remain fixed throughout sampling, preventing the
conditioning evidence from being corrupted or implicitly reconstructed by the
ODE.

\paragraph{Reference tokens.}
Geometry encoders are set-based, so encoding a reference jointly with unavailable
target views would leak target-set context during training and create a
train--test mismatch. We instead encode the $K$ observed references jointly,
but using only those references as the encoder context available at inference:
\begin{equation}
    \bar{\mathbf{z}}^{\mathrm{ref}}=
    \operatorname{Std}\!\left[
    \mathrm{Enc}_{\phi}^{\mu}\!\left[
    \mathcal{P}\!\left(\mathcal{E}(\mathbf{I}^{\mathrm{ref}}_{1:K})\right)
    \right]\right].
    \label{eq:z_ref}
\end{equation}
The resulting tokens preserve the same geometry-native representation used for
the generated views, allowing evidence and state tokens to interact directly in
self-attention. They are assigned $t=0$, prepended to the noisy all-view tokens,
and excluded from the prediction head. Consequently, their role is purely
contextual: they can be attended to at every transformer layer but are never
updated by the numerical solver. Every output slot---including a requested
output at a reference camera---remains an ordinary noisy flow variable. This
separation avoids repeatedly clamping a partially denoised reference slot and
keeps the same ODE parameterization for conditional and unconditional sampling.

\paragraph{Camera rays.}
For each latent cell we form a world-space Pl\"ucker ray $(\mathbf{d},\mathbf{m})$,
where $\mathbf{m}=\mathbf{o}\times\mathbf{d}$, and decompose it as
\begin{equation}
    \mathbf{d},\qquad
    \hat{\mathbf{m}}=\frac{\mathbf{m}}{\|\mathbf{m}\|+\epsilon},\qquad
    s=\log(\|\mathbf{m}\|+\epsilon).
    \label{eq:plucker_decomp}
\end{equation}
Here $\mathbf{d}$ describes viewing direction, $\hat{\mathbf{m}}$ captures the
orientation of the ray moment, and the log-scale channel $s$ retains metric
translation magnitude when metric camera sidecars are available. Following
SCoPE~\citep{scope2026}, the resulting embedding modulates query/key
self-attention, making pairwise token interaction explicitly dependent on the
corresponding camera rays rather than treating pose as a single global vector.
Rays are evaluated at the DiT~\citep{peebles2023scalable} token grid, preserving pixel--ray alignment when
the latent patch size changes. Reference tokens receive the rays of their
observed cameras, while noisy target tokens receive rays from the requested
output cameras; the same spatial representation therefore relates evidence to
the views being generated.

\paragraph{Text.}
Prompts are encoded by a frozen language model and projected to the transformer
cross-attention width. Unlike reference and camera signals, which are aligned
with latent locations and enter self-attention, text acts as global semantic
context through cross-attention. Independent dropout of text, reference, and
camera conditions trains a single set of weights on the subsets of
$\mathcal{C}$ encountered at inference. The same model can consequently operate
in text-only, camera-controlled, reference-conditioned, or jointly conditioned
image and video generation regimes, and the dropped-condition branches also
provide the unconditional predictions used for classifier-free guidance~\citep{ho2022classifier}.

\subsection{Two-stage training, sampling, and evaluation protocols}

We distinguish two training configurations: a controlled setting for comparing
latent representations and a separate setting for the final model used in
qualitative demonstrations. Evaluation covers three complementary aspects:
codec reconstruction, conditional generation, and native geometry readout.

\paragraph{Two-stage training and shared sampling procedure.}
Stage~1 jointly trains $\mathcal{A}_{\phi}$ and
$\mathcal{D}_{\theta}^{\mathrm{rgb}}$ on mixed single- and multi-view data,
with $\mathcal{E}$ and $\mathcal{H}_{\mathrm{DPT}}$ frozen.
Stage~2 freezes all Stage~1 modules and trains only the conditional flow model.
Both configurations use the same sampling procedure:
reference images are encoded once as clean conditions, while all requested
output slots are initialized with Gaussian noise. We jointly integrate their
velocity field from $t=1$ to $t=0$, then denormalize and decode the resulting
latents through Eq.~\ref{eq:ngd_readout}. Quantitative comparisons and
qualitative demonstrations both use 50 Euler steps and a CFG scale of $2$.
For each controlled backend, the latent normalization statistics used
during training are also applied at inference.

\paragraph{Controlled comparison: architecture and training data.}
To isolate the effect of latent representation, we place all compared
backends within the same DDT-style flow architecture and match their
training and sampling protocols. The shared transformer comprises
28 encoder blocks of width 768 and six decoder blocks of width 2048.
It predicts clean latents, uses an eight-block auxiliary base head, and
incorporates frozen Qwen3-0.6B~\citep{yang2025qwen3} text features through
cross-attention. The shared flow backbone contains approximately $0.93$B
parameters; total parameter counts differ slightly because each latent
shape requires its own input and output projections.

Training clips contain nine $252{\times}252$ views drawn from an equal
mixture of RealEstate10K~\citep{zhou2018re10k} and
DL3DV~\citep{ling2024dl3dv}, with one to four prefix views serving as
references. Each model is trained on eight GPUs.
The controlled comparison therefore holds the flow backbone, data mixture,
optimization schedule, and sampling protocol fixed while varying the latent
encoder, decoder, and the projections required to interface with them.
Architecture- or conditioning-specific modifications are introduced only
in the corresponding ablations.

\paragraph{Controlled comparison: optimization.}
We train in BF16 using AdamW with $\beta=(0.9,0.95)$, zero weight decay,
gradient clipping at $1.0$, and an EMA decay of $0.9995$.
The learning rate increases to $10^{-4}$ over 2k warmup steps and then
follows a cosine schedule toward $10^{-6}$ over a 100k-step horizon.
Text, camera, and reference conditions are independently dropped with
probabilities $0.10$, $0.05$, and $0.05$, respectively.
A single-view text-to-image batch is interleaved every three multi-view
updates.

\paragraph{Codec evaluation.}
We first evaluate the latent representations through direct codec readout,
independently of flow-model generation.
The latent-space study in Section~\ref{sec:exp_latent} uses
ImageNet-S~\citep{gao2022large} for single-view diagnostics.
Its multi-view diagnostics use video clips from five sources:
RealEstate10K~\citep{zhou2018re10k},
DL3DV~\citep{ling2024dl3dv},
MVS-Synth~\citep{huang2018deepmvs},
ScanNet++~\citep{yeshwanth2023scannet++}, and
Open-Sora Plan (OSP)~\citep{lin2024open}.
The RGB reconstruction study in Section~\ref{sec:exp_recon} uses
ImageNet for single-view reconstruction and reports the average
multi-view reconstruction performance over the same five video sources.
These codec evaluations are separate from the held-out scene pools used
for the generation comparison below.

\paragraph{Generation evaluation and ranking.}
The main generation comparison uses shared held-out pools of 64 scenes
from each of RealEstate10K and DL3DV, totaling 128 scenes.
Each example contains nine views at $252{\times}252$, with one clean
reference view and a prescribed camera trajectory.
The frame interval is 10 for RealEstate10K and 1 for DL3DV.
All backends use these same examples and the shared sampling procedure
described above.

SD-VAE~\citep{rombach2022ldm},
WAN2.1 VAE~\citep{wan2025wan}, and
RAEV2~\citep{raev22026} produce $256{\times}256$ images in this setting.
We resize their generated outputs to $252{\times}252$ before evaluation
to match the other backends and ground-truth clips.
FID and FVD measure frame- and clip-level distributional quality,
respectively, whereas LPIPS, PSNR, and SSIM compare generated frames with
their paired targets along the prescribed trajectory.
Blue cells rank only the seven latent variants trained with the shared
flow recipe. GLD and Gen3R are reported as external system-level
references and are excluded from this controlled ranking.

\paragraph{Camera accuracy and cross-view consistency.}
To evaluate camera consistency, we apply the same estimator, VGGT or
DA3-GIANT, to each generated nine-view clip and its corresponding real video.
We compare the two recovered camera trajectories after Umeyama Sim(3)
alignment, then compute absolute trajectory error (ATE), relative
translation error (RPEt), and relative rotation error (RPEr).
VGGT is independent of all evaluated latent backbones and serves as
the independent evaluator; DA3-GIANT provides an additional
cross-check with backbone overlap.

For reprojection evaluation~\citep{gld2026}, we use the estimator's recovered
world points, cameras, and intrinsics. High-confidence points are transferred
between views and projected back to the source view. The resulting round-trip
2D pixel error is normalized
by the image diagonal. We report Reproj@50, retaining the top half of
points by estimator confidence.

We compute MEt3R using its official
implementation~\citep{asim2025met3r}. For each pair of generated views,
it recovers pairwise geometry, upsamples visual features with FeatUp,
and renders the features into the opposite view using visibility and
z-buffer occlusion masks. The score measures symmetric cosine-feature
disagreement and is averaged over all $\binom{V}{2}$ view pairs.
Both Reproj and MEt3R are lower-is-better metrics.

\paragraph{Native geometry evaluation.}
Complementing the RGB-based evaluation above, we directly assess the
geometry decoded from the latent representation.
Decoded depth and point maps are compared with reference predictions
obtained by applying Pi3~\citep{wang2026pi3} to the real target frames.
Depth is aligned using a per-view median scale, while point clouds use
the Sim(3) transformation estimated from the camera trajectories.
Native camera ATE and RPEr are evaluated against the ground-truth
dataset trajectory.
This protocol evaluates the model's own geometry readout, separately
from the post-hoc reconstruction of generated RGB frames used for
camera and consistency evaluation.

\paragraph{Raw L0 readout and special-token initialization.}
The Raw geometry (DA3-GIANT L0) baseline retains spatial patch tokens,
rather than the complete intermediate state of the DA3 backbone. To obtain
its geometry readout, we propagate these patch tokens through the remaining
frozen backbone and pass the resulting feature hierarchy to the frozen DPT
head. The non-spatial, image-dependent camera/CLS token states are not part
of the generated patch-token grid. Modeling these states jointly with the
spatial representation would require an additional generative interface;
we therefore initialize the missing special-token states to zero and use
this same convention for reconstruction and generation. This is a modeling
choice, not a claim that special tokens are inherently impossible to
generate. In particular, supplying special tokens extracted from real
target images during generation would introduce information unavailable
at inference.

For reconstruction, the real images are available, so their original
special-token states provide a diagnostic control. Table~\ref{tab:raw_l0_cls}
compares correct-CLS and zero-CLS readouts on the same 64 DL3DV scenes,
holding the L0 patch features fixed and varying only the special-token
input. Zero initialization has a small effect on depth accuracy and a
more noticeable effect on camera-aligned point-cloud and pose errors.
Consequently, the reported Raw L0 reconstruction is a patch-only,
generation-compatible readout, not the original full DA3 forward pass.
If all required intermediate states are preserved and the frozen backbone
is resumed under identical settings, it should reproduce the original
features and predictions. The correct-CLS control below is not, by itself,
a verification of that full-forward equivalence; moreover, the camera
poses in this evaluation are recovered from decoded ray maps rather than
DA3's native camera head. The comparison therefore assesses geometry
retention under the stated latent interface and should not be interpreted
as showing that the compact codec outperforms the full DA3 model.

\begin{table}[!ht]
\centering
\caption{\textbf{Special-token control for Raw L0 reconstruction.}
The same 64 DL3DV scenes and L0 patch features are used in both rows.
Depth and point maps use Pi3 on the real images as the reference; poses
use the dataset cameras, with the alignment protocol described above.
Correct CLS uses original special-token states from the real inputs
and is a reconstruction-only control. Zero CLS is the setting used
for the patch-only baseline.}
\label{tab:raw_l0_cls}
\small
\setlength{\tabcolsep}{4pt}
\begin{tabular}{lcccccc}
\toprule
Special-token input & AbsRel$\downarrow$ & $\delta_1\uparrow$
& Cham.$\downarrow$ & PMap$\downarrow$ & ATE$\downarrow$
& RPEr$^{\circ}\downarrow$ \\
\midrule
Correct CLS & 0.1265 & 0.8604 & 0.1475 & 0.4383 & 0.00779 & 0.2233 \\
Zero CLS    & 0.1274 & 0.8581 & 0.1824 & 0.4826 & 0.00873 & 0.2541 \\
\bottomrule
\end{tabular}
\end{table}

\paragraph{Final model for qualitative demonstrations.}
The final model is trained separately for longer, higher-resolution
generation and is excluded from the controlled ranking.
It uses a frozen space--time codec built on DA3-GIANT features at
$672{\times}378$ and trains the flow model on 81-view clips with one to
four prefix reference views.
Its multi-domain training mixture includes
RealEstate10K~\citep{zhou2018re10k},
DL3DV~\citep{ling2024dl3dv},
ScanNet++~\citep{yeshwanth2023scannet++},
MVS-Synth~\citep{huang2018deepmvs},
SpatialVID~\citep{wang2026spatialvid},
OmniWorld~\citep{zhou2025omniworld},
training data curated and released by
Open-Sora Plan~\citep{lin2024open}, and several internally curated datasets.

We train the $0.93$B-parameter flow transformer on 80 GPUs in BF16 using
AdamW, with 500 warmup steps to a base learning rate of $5{\times}10^{-5}$,
followed by cosine decay to $10^{-6}$ by 60k steps.
Gradient clipping is set to $1.0$ and EMA decay to $0.9995$.
The parameter count refers only to the trainable Stage~2 generator and
excludes the frozen DA3 encoder, space--time codec, and text encoder.
Text, camera, and reference conditions are independently dropped with
probabilities $0.10$, $0.20$, and $0.20$, respectively.
Single-view text-to-image batches at the same aspect ratio are
interleaved every four multi-view updates.
We use shifted logit-normal time sampling, with a shift of $6.364$ for
81-view training and $4.5$ for the single-view stream.
For the showcased results, all 81 requested views are sampled jointly
from one reference using the shared sampling procedure above.
Camera conditioning is retained in the unconditional CFG branch.

\section{Latent-space diagnostic definitions}
\label{app:diagnostics}
\label{app:metric_defs}

Table~\ref{tab:latent_diagnostics} characterizes seven complementary properties
of a latent representation: transport ambiguity ($\rho$), covariance
conditioning ($\kappa$ and effective rank), semantic neighborhoods (LNC@5),
within-image spatial structure (LDS and SRSS), and cross-view correspondence
(xLNC$^{*}$). The transport and semantic-neighborhood perspectives follow the
latent-diffusability study of \citet{diffusingrightspace2026}, while LDS and
SRSS follow the spatial diagnostics introduced and released by
\citet{singh2025matters}. The covariance quantities are standard spectral
statistics. We introduce xLNC$^{*}$ as a multi-view extension of
latent-neighbor consistency, using dense correspondences to define cross-view
positives.

\paragraph{Evaluation data and preprocessing.}
For the single-view columns, we use ImageNet-S~\citep{gao2022large} at
resolution $256$, which provides ImageNet class labels and pixel-level
foreground masks. For the multi-view columns~\cite{zhou2018re10k,ling2024dl3dv,huang2018deepmvs,yeshwanth2023scannet++,wang2026spatialvid}, each short scene clip defines one
group, so views of the same scene determine the within-group variation. We map
spatial neighborhoods to image coordinates before sampling token pairs, making
the distance thresholds comparable across latent grids. Tokens are
$\ell_2$-normalized before cosine-based measurements. The table reports the raw
representation consumed by the flow model, without diagnostic-only PCA removal
or feature centering.

\paragraph{Within-group velocity fraction $\rho$ (lower is better).}
Let $\boldsymbol{\Sigma}_{\mathrm{within}}$ denote the covariance of latent
variation within a semantic class for single-view evaluation, or within a scene
clip for multi-view evaluation, and let $\boldsymbol{\Sigma}_{\mathrm{total}}$
be the covariance over the complete evaluation set. We compute
\begin{equation}
    \rho=
    \frac{\operatorname{tr}(\boldsymbol{\Sigma}_{\mathrm{within}})}
         {\operatorname{tr}(\boldsymbol{\Sigma}_{\mathrm{total}})}.
\end{equation}
This normalized variance ratio accompanies the velocity-ambiguity analysis of
\citet{diffusingrightspace2026}. A low value indicates that samples sharing a
class or scene occupy a compact portion of the overall latent distribution and
therefore induce less ambiguous transport. Because both terms are traces, the
ratio is dimensionless and is not artificially improved by appending inactive
channels.

\paragraph{Condition number $\kappa$ and effective rank (lower/higher is
better).}
We flatten the spatial positions and estimate the channel covariance of the
latent. If its eigenvalues are $\lambda_1\geq\cdots\geq\lambda_C$, then
\begin{equation}
    \kappa=\frac{\lambda_1}{\lambda_C},
    \qquad
    \operatorname{effrank}=
    \frac{\left(\sum_i\lambda_i\right)^2}
         {\sum_i\lambda_i^2}.
\end{equation}
$\kappa$ is the covariance condition number; a large value indicates severe
anisotropy or near-rank deficiency. Effective rank is the spectral participation
ratio. It equals $r$ for a perfectly flat rank-$r$ spectrum but remains a
continuous concentration statistic for realistic spectra. Reporting both
quantities distinguishes nominal channel count from the number of directions
that carry substantial variance.

\paragraph{Latent neighbor consistency LNC@5 (higher is better).}
Following the semantic-separability diagnostic of
\citet{diffusingrightspace2026}, we foreground-pool each ImageNet-S latent into
one descriptor. We retrieve its five nearest descriptors using cosine
similarity and average the fraction whose ImageNet label matches the query.
LNC@5 therefore measures local semantic organization; it does not by itself
measure reconstruction fidelity or patch-level geometry.

\paragraph{Latent distance structure LDS (higher is better).}
LDS measures whether nearby image locations remain more similar than distant
locations in latent space. For each image, we average token cosine similarity
over spatially local pairs and subtract the average over far pairs:
\begin{equation}
    \mathrm{LDS}=
    \mathbb{E}_{(i,j)\in\mathcal{N}}[\langle\hat{\mathbf z}_i,
    \hat{\mathbf z}_j\rangle]
    -
    \mathbb{E}_{(i,j)\in\mathcal{F}}[\langle\hat{\mathbf z}_i,
    \hat{\mathbf z}_j\rangle].
\end{equation}
We follow the released LDS implementation of \citet{raev22026,singh2025matters}. Its local
and far neighborhoods are expressed in common image coordinates (approximately
$24$ and $96$ pixels in our evaluation), so the comparison is not tied to a
particular token-grid resolution. A high score means that latent similarity
decays meaningfully with spatial distance instead of collapsing to a nearly
uniform descriptor.

\paragraph{Spatial relative similarity score SRSS (higher is better).}
SRSS uses the ImageNet-S foreground mask to test region-level organization. For
each foreground anchor, it compares similarity to nearby foreground tokens with
similarity to distant background tokens:
\begin{equation}
    \mathrm{SRSS}=
    \mathbb{E}_{i\in\mathrm{FG}}
    \left[
    \mathbb{E}_{j\in\mathcal{P}(i)}
       \langle\hat{\mathbf z}_i,\hat{\mathbf z}_j\rangle
    -
    \mathbb{E}_{k\in\mathcal{B}(i)}
       \langle\hat{\mathbf z}_i,\hat{\mathbf z}_k\rangle
    \right],
\end{equation}
where $\mathcal{P}(i)$ contains nearby foreground positives and
$\mathcal{B}(i)$ contains spatially distant background negatives, as defined by
the ImageNet-S masks. We follow the released SRSS implementation of
\citet{zhong2026diffusing,singh2025matters}. A high score indicates that foreground regions are
locally coherent while remaining distinguishable from background content.

\paragraph{Cross-view latent neighbor consistency xLNC$^{*}$ (higher is
better).}
This is our multi-view extension of LNC. We first use the UFM-Refine dense
correspondence model~\citep{zhang2025ufm} to establish matches between two
views. For every matched query token,
we retrieve its nearest token in the other view by latent cosine similarity and
count the retrieval as correct when it falls within the matched target
neighborhood. We chance-normalize the resulting accuracy,
\begin{equation}
    \mathrm{xLNC}^{*}=
    \frac{a-a_{\mathrm{chance}}}{1-a_{\mathrm{chance}}},
\end{equation}
where $a$ is the measured retrieval accuracy. Thus, 0 denotes chance-level
retrieval and 1 denotes perfect correspondence. The main table evaluates view
pairs with sufficient visual overlap. Unlike LDS and SRSS, which characterize
within-image organization, xLNC$^{*}$ directly measures whether a latent
retrieves the same scene content across changes in viewpoint.

\section{Additional qualitative comparisons}
\label{app:qualitative}

\subsection{Independent 3D check}

\begin{figure}[!htp]
\centering
\includegraphics[width=\linewidth]{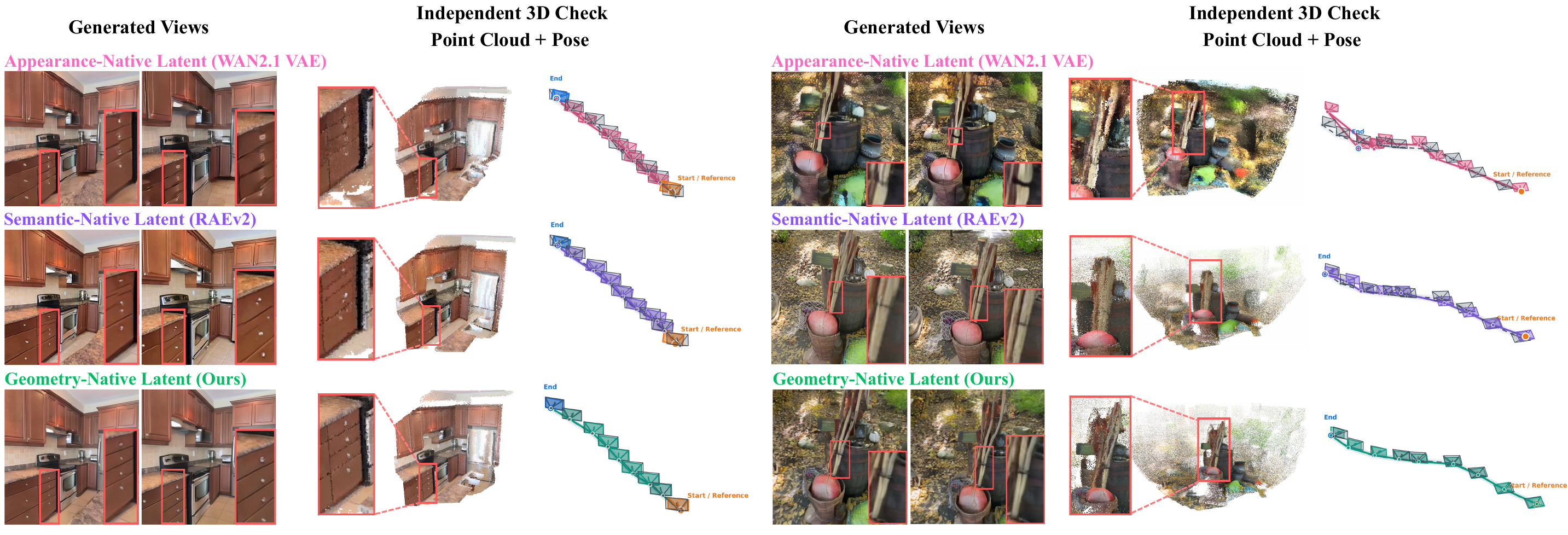}
\vspace{-2em}
\caption{\textbf{Independent 3D check of generated views.}
The appearance-native WAN2.1 VAE, semantic-native RAEV2, and our
geometry-native latent receive the same reference and camera controls. The
right-hand panels show point clouds and camera trajectories reconstructed from
the generated RGB clips by an independent geometry model VGGT.}
\label{fig:app_native_geometry}
\end{figure}

Figure~\ref{fig:app_native_geometry} compares appearance-native,
semantic-native, and geometry-native generated states on two scenes. Alongside
the generated views, an independent reconstruction pipeline recovers a point
cloud and camera trajectory from each output clip. This separates visual
plausibility from whether the generated frames jointly support a coherent 3D
scene and camera path.

\subsection{RGB comparisons}

\begin{figure}[!htp]
\centering
\hspace*{-1cm}\includegraphics[width=\linewidth]{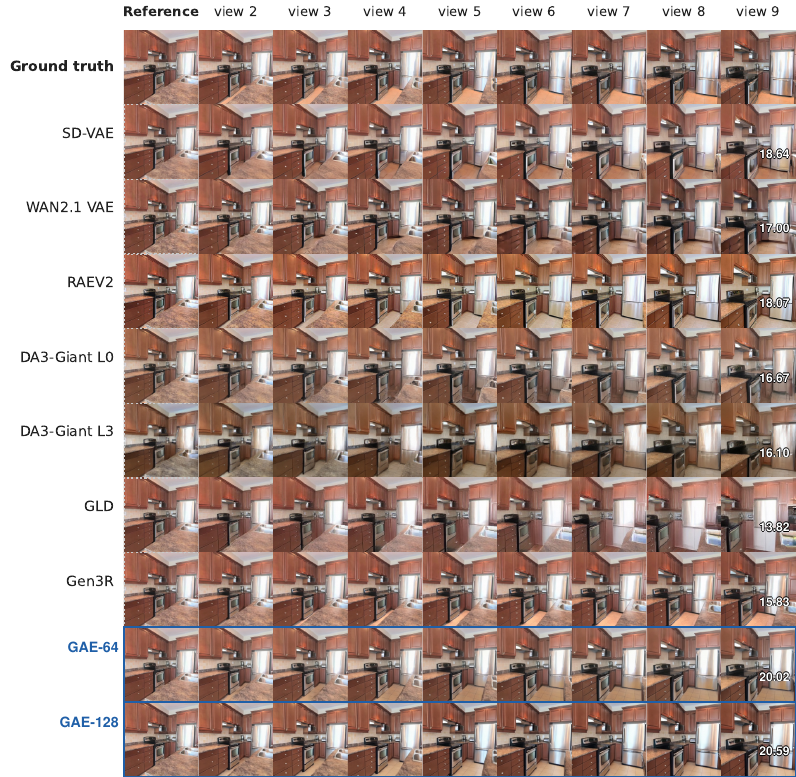}
\caption{\textbf{RealEstate10K comparison.} Given the same reference image and
target camera trajectory, all nine requested views are generated jointly,
including the output at the reference camera pose.}
\label{fig:qual_rgb_re10k_full}
\end{figure}

\begin{figure}[!htp]
\centering
\hspace*{-1cm}\includegraphics[width=\linewidth]{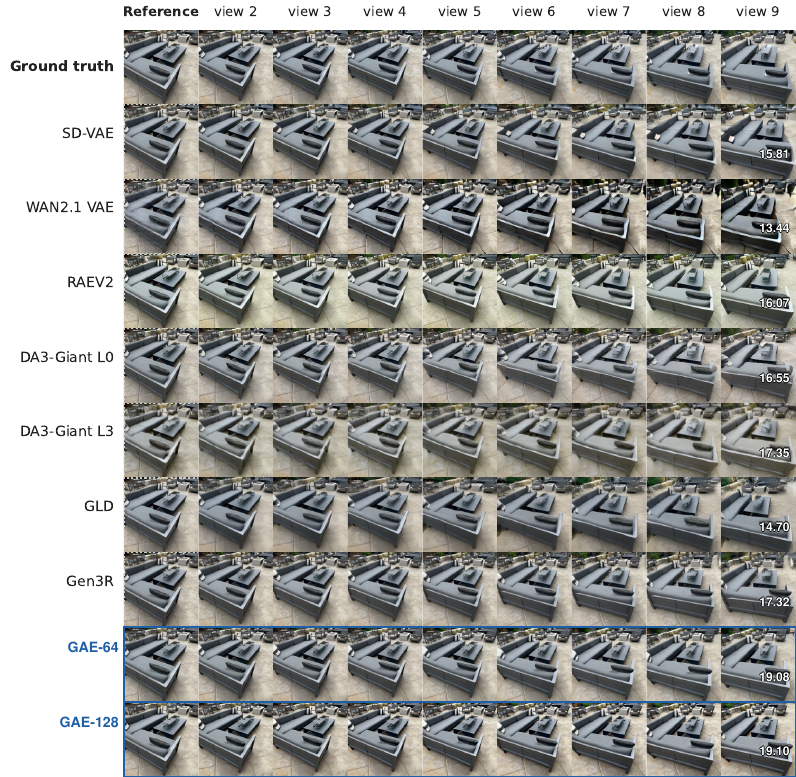}
\caption{\textbf{DL3DV comparison.} Given the same reference image and
target camera trajectory, all nine requested views are generated jointly,
including the output at the reference camera pose.}
\label{fig:qual_rgb_dl3dv_full}
\end{figure}

Figures~\ref{fig:qual_rgb_re10k_full} and~\ref{fig:qual_rgb_dl3dv_full} show all
nine views for one RealEstate10K scene and one DL3DV scene from the shared
$64$-scene evaluation pools. The displayed PSNR is computed over the eight generated
views of that scene, not averaged across the dataset.

\end{document}